# Exploring Brain-wide Development of Inhibition through Deep Learning

Asim Iqbal[1,2], Asfandyar Sheikh[3], Theofanis Karayannis[1,2]

[1]Laboratory of Neural Circuit Assembly, Brain Research Institute (HiFo), UZH
[2]Neuroscience Center Zurich (ZNZ), UZH/ETH Zurich
[3]Department of Information Technology and Electrical Engineering (D-ITET), ETH Zurich

*Correspondence should be addressed to T.K. (karayannis@hifo.uzh.ch)

## Abstract

We introduce here a fully automated convolutional neural network-based method for brain image processing to **De**tect **N**eurons in different brain **R**egions during **D**evelopment (*DeNeRD*). Our method takes a developing mouse brain as input and i) registers the brain sections against a developing mouse reference atlas, ii) detects various types of neurons, and iii) quantifies the neural density in many unique brain regions at different postnatal (P) time points. Our method is invariant to the shape, size and expression of neurons and by using *DeNeRD*, we compare the brain-wide neural density of all GABAergic neurons in developing brains of ages P4, P14 and P56. We discover and report 6 different clusters of regions in the mouse brain in which GABAergic neurons develop in a differential manner from early age (P4) to adulthood (P56). These clusters reveal key steps of GABAergic cell development that seem to track with the functional development of diverse brain regions as the mouse transitions from a passive receiver of sensory information (<P14) to an active seeker (>P14).

With each passing day in the field of neuroscience a new dataset is collected that needs to go through a rigorous procedure of data convertibility, data compatibility, and data analysis. The term 'big data' is now commonly used in neuroscience consortiums since biology labs across the world are collecting terabytes of neuroimaging data everyday. Neuroscientists are in need to explore these datasets to unravel hidden patterns in human and non-human brain tissues. To further deepen our understanding of the structure and function of the nervous system, there are a series of technological advancements required, both at the hardware and software end. There has been a tremendous advancement of hardware development in the last couple of decades to record the activity of thousands of brain cells at a time through multichannel recording probes [1] or large field of view fluorescent microscopes [2], although the cells are usually confined to a given brain region. At the same time, big consortia and prime institutes world-wide e.g., Allen Brain Institute [3], Blue Brain [4], and Human Brain Projects [5] are producing high-throughput genetic and structural imaging data that capture and visualize many different types of neurons in the whole brain with spatially precise single-cell-resolution. Although efforts have been made in counting all the cells in a given brain region, these methods rely on a specific areal dissection, they are not easily adaptable to detecting specific cell types and lose crucial information on the position of cells in the tissue along the dorso-ventral, antero-posterior, and medio-dorsal axes [6]. Nevertheless, recent advances in methods to identify individual cells in defined regions have provided a powerful tool to perform high-throughput analysis of cell numbers *in situ* [7, 18], but these methods are limited in scope due to a large variation of size, shape, and density of neurons in different regions of the brain.

We introduce here a Deep Learning method, based on the Faster Regions with Convolutional Neural Network (RCNN) [8], which we call *DeNeRD* as shown in **figure 1 (a)** to expedite brain-wide data analysis by detecting neurons in brain sections *in situ*. *DeNeRD* takes images of mouse brain sections as input and with some pre-processing, it forwards them to the *Neuron Detector* and *Brain Registrator* modules, which detect the neuronal population in any given brain section and register it with Allen Institute's developing mouse brain reference atlas [3]. Our method is invariant to shape, size, density and expression of neurons in single (**figure 1 (b) and (c)**) as well as different brain areas (**supplementary figure 1**) and outperforms the commonly used cell detection techniques [9] (**supplementary figure 2 (a-b)** and **supplementary figure 3**). *DeNeRD* is applicable on brain sections available at Allen Brain Institute's online public repository (OPR) [3] and is scalable to detect neurons in brain images of different kinds that it has never "seen" before by generating the respective ground truth data through our Simple Graphical User Interface (*SiGUI*) Software (**supplementary figure 10**).

Here we utilize our method to understand brain-wide development of inhibition. A large population of neurons in the brain, including the second largest in the cortex, consists of GABAergic neurons. These neurons are inhibitory in nature and have therefore been shown to regulate a number of processes in the brain to allow an increased capacity of information processing. As their name implies, they utilize GABA as a neurotransmitter, which they produce using an enzyme called Glutamate Decarboxylase (GAD), which they pack in vesicles for release using the Vesicular GABA Transporter (VGAT). There are two isoforms of GAD, 1 (GAD67) and 2 (GAD65), and although their subcellular location is considered different, they are thought to be expressed in all GABAergic neurons.

We hypothesized that the amount of GABAergic neurons a brain region holds and the functional maturation of those neurons is in direct correlation to the functional maturation of that brain region. We therefore utilized the GAD1 marker as a proxy for neuronal number/density of GABAergic neurons and, VGAT for capturing the maturation of the output of GABAergic neurons. VGAT labels the subset of neurons in development that have acquired the capacity to pack GABA in synaptic vesicles and hence release it. Analyzing the GAD1 marker at three developmental stages, P4, P14, and P56 using *DeNeRD* allowed us to identify novel developmental motifs in the cell death of GABAergic neurons throughout the brain. The development of the VGAT marker on the other hand pointed at the distinct functional maturation of different regions of the brain, as indicated by the increased need for inhibitory control.

We report at least 6 unique clusters in the brain (**figure 2**) in which GABAergic neural density (GAD1) undergoes a series of alterations during the postnatal life of mouse. As expected, the output of these GABAergic neurons (VGAT) in most of the brain regions seems to have a direct relation with the activation of sensory modalities of the animal, especially when the mouse transitions from only a handful of active sensory modalities being active (e.g., touch and smell at ~P4) to actively exploring the environment using vision and auditory modalities (~P14). In addition, a number of interesting observations, previously not possible with other methods, are discussed. Following is a description of each cluster in **figure 2 (a-f)**. The color-coded sample brain regions are shown in **figure 2 (g)** and the neural population of each region can be found in **supplementary figure 19**.

**Cluster#1: P4≅P14≥P56.** Thalamus, rhombomeres: R3-R4, R5-R6, R7-R11 and spinal cord (SC) regions are found in this cluster (**figure 2 (a)**). These are structures that differentiate early on in development and this is reflected in the density of GABAergic interneurons (GAD1), which do not change from P4 to P14. However, the analysis points at a decrement in neural density after P14 ($p \leq 0.05$, $p \leq 0.05$, $p \leq 0.01$ and $p \leq 0.01$), which suggests a late neural apoptosis, much further in development than previously reported in the brain [10]. As would be expected by previous reports, the thalamus gets active and mature by P4 (VGAT), and although there is a trend towards reduction in activity and maturity between P14 and P56 to match the decrease in neural density, there is no significant change. However, a significant decrement in activity is observed for rhombomeres (R3-R4, R5-R6) and spinal cord from P14 to P56 density ($p \leq 0.05$, $p \leq 0.01$ and $p \leq 0.01$).

**Cluster#2: P4≤P14≥P56.** The Isocortex is found in this cluster (**figure 2 (b)**) with an increment in neural density of GABAergic neurons from P4 to P14 ($p \leq 0.05$). Since there is no GABAergic neurogenesis occurring in postnatal stages, the analysis indicates that GAD1 expression is not complete at P4. Intriguingly, though we find that there is a decrement in neural density from P14 to P56 ($p \leq 0.05$), this is in contrast to previous reports showing that the majority of cell death occurs before the end of the second post-natal week in this brain structure. Activity of GABAergic neurons is significantly increased from P4 to P14 ($p \leq 0.001$), which is also a critical time window [11] when most of the sensory modalities (opening of eyes, auditory canal and motor activity [12] etc.) are getting active. Similar to the neural density ($p \leq 0.05$), the activity of GABAergic neurons (VGAT) in isocortex also decreased substantially ($p \leq 0.001$) after P14.

**Cluster#3: P4≥P14≅P56.** Three brain regions are discovered in this cluster: Midbrain, Tectum and Preoptic area (**figure 2 (c)**). In the Midbrain region, GABAergic neural density is all set by P4, but only about half of them are making active synaptic connections. Since the functional role of the midbrain is also to modulate motor activity, it is not surprising that GABAergic neurons are not completely active around P4 ($p \leq 0.001$). However, as the mouse starts to explore its environment (P14), the activity of GABAergic neurons (VGAT) is doubled and matches the GABAergic neural density (GAD1). A somewhat similar trend is also observed in the Tectum and Preoptic area, with the former being involved in controlling eye and auditory reflexes [13], whereas the latter regulates thermoregulation [14].

**Cluster#4: P4≥P14≤P56.** The Hippocampus undergoes a decrease in neural density of GABAergic neurons from P4 to P14 ($p \leq 0.001$) but an increase from P14 to P56 ($p \leq 0.01$) (**figure 2 (d)**). VGAT, on the other hand, is increased from P4 to P14, suggesting an enhancement of GABAergic cell output, but is decreased from P14 to P56. Although this may seem like a paradox since one would expect that GAD1 and VGAT would track a similar neural density in the adult, this discrepancy is due to the signal being affected by the development of the GAD1 signal in the granule cells of the dentate gyrus (DG). It is known that these cells can co-release glutamate and GABA early in development and at P14, but only glutamate when they are about 3 weeks [15]. Our data reveals that although these cells produce GABA in the adult, they cannot pack it into vesicles since VGAT is not there and hence loose the capacity to release it.

**Cluster#5: P4≅P14≤P56.** There is observed a slight increase in neural density of GABAergic neurons from P14 to P56 ($p \leq 0.05$) in Olfactory bulb (+ Striatum) region which is due to adult neurogenesis occurring in the olfactory bulb region (**figure 2 (e)**) as reported in the previous studies [16]. However, an increase in activity of GABAergic neurons is observed from P4 to P14 ($p \leq 0.001$) but it gets stable at P14 and does not undergo any significant change in the adult mice (P56).

**Cluster#6: P4≅P14≅P56.** Six brain regions are found in this cluster where GABAergic neural density does not go through any significant change from P4 to P56 (**figure 2 (f)**). Pre-thalamic and hypothalamic regions are found herein, along with the developing hindbrain region. A significant increase in activity of GABAergic neurons (VGAT) is found in most of these regions from P4 to P14, followed by a stability in neural density and functional activity until the P56 time point.

The data presented above indicates that even though we see cell death in many regions from P4 to P14, as expected, there are two clusters that include the Isocortex and the Thalamus where we see a reduction in cell numbers between P14 and P56. At the same time, there are regions that seem to mature by P4, since there is no change in cell numbers from P4 to P14 (cluster # 1, 5 and 6).

Even more interesting observations come up by looking at the neural density of GAD1 versus VGAT. One would expect that the latter always trails behind the former between P4 and P14, which should ultimately catch up at P14 or at least P56. Although this is in general the case, there are 5 regions in the brain where the number of neurons expressing VGAT are less than the ones expressing GAD1 at P56. These are the Isocortex, Midbrain, Hippocampus, Olfactory bulb (+Striatum) and Hindbrain regions. If one looks carefully at the signals in the Hippocampus, it becomes clear that this difference is due to the expression of the two markers in the DG granule cells. Even though this may seem odd, it fits well with the previously published findings, which show that these cells can co-release glutamate and GABA early on in development, but only glutamate in the adult brain [12]. Our findings suggest that even though adult granule cells produce it, they can no longer pack it into vesicles and hence release GABA as they could at P14. These findings not only demonstrate the power of *DeNeRD* to produce results in line with previous studies, but to also point to novel hypotheses that similar co-release mechanisms may occur in parts of the Isocortex and Striatum. In addition, the Midbrain shows the same pattern, even though unlike the Hippocampus, it is a site where cells that can co-release glutamate and GABA are found to be present in the adult brain [17]. Nevertheless, at P14 there is no difference in the cell numbers between GAD1 and VGAT, suggesting that all designated cells can co-release the two neurotransmitters at that time point, including the ones that cannot do so in the adult, including those projecting to the Nucleus Accumbens (NAc) and Ventral Pallidum (VP). The Hindbrain is the only one of the 4 structures where VGAT never catches up with GAD1 and this may be due to the molecular layer interneuron expression patterns, but this requires further research.

In conclusion, using the power of *DeNeRD* we are able to point towards potentially remarkable differences in the cell death of GABAergic neurons in different brain regions during development, as well as towards some very intriguing observations on the functional maturation of the brain, which would have otherwise been masked under the complexity and scale of the data.

**Acknowledgements**
We thank A. Ö. Argunsah (Brain Research Institute, University of Zurich) for providing valuable feedback and we also thank R. Khan (Department of Biology, ETH Zurich) for helping us with the brain registrator pipeline.

**Author contributions**
A.I. and T.K. conceptualized the study and wrote the paper. A.I and A.S developed the neural detector pipeline and performed the testing comparisons of *DeNeRD* with other computational cell-counting methods. A.I. developed the brain registrator pipeline. A.S. developed the *SiGUI* software for generating ground-truth data. A.I. performed all the computational and statistical analysis on GAD1 and VGAT developing brains.

**Competing interests**
The authors declare no financially competing interests.

# Methods

**Developing mouse brain sections.** Brain sections of standard ISH quality were collected from Allen brain's OPR. The developing mouse brains were chosen at three unique post-natal time points: P4, P14 and P56 and for two different markers: GAD1 (Glutamate decarboxylase 1) and VGAT (GABA vesicular transporter). Along with the ISH brain sections, expression sections of the same brains were also imported, which were utilized for ground-truth dataset generation for Faster R-CNN, explained in ground-truth labeling section. These ISH brain sections are $20\mu m$ thick, sectioned at $200\mu m$ apart – covering an entire hemisphere from lateral to medial direction. Some lateral and medial examples from adult mouse brain are shown in (**supplementary figure 4 (b)**) and (**supplementary figure 5 (b)**) respectively. In total, six brains were analyzed – a pair of GAD1 and VGAT brains at each developmental time point.

**Brain registrator pipeline.** For the purpose of brain registration, standard Nissl sagittal sections of P4, P14 and P56 mouse brains are also collected from Allen brain's OPR. As an example, lateral and medial Nissl sections of an adult brain are shown in (**supplementary figure 4 (c)**) and (**supplementary figure 5 (c)**), respectively. Allen developing mouse brain reference atlases are pre-registered with these Nissl sections as shown in (**supplementary figure 4 (e)**) and (**supplementary figure 5 (e)**).

Affine algorithm in elastix toolbox [19] is optimized for the purpose of registration of brain sections. The affine transformation takes a fixed and moving image as input and returns a transform parameter map which can be applied to the moving image. The un-registered brain section ($B$) with the corresponding registered Nissl section ($N$) are passed to the pipeline as fixed and moving image. $B$ and $N$ are first down-sampled by a reasonable ratio ($r$) such as to reduce the computational cost of brain registrator pipeline without compromising the image quality of brain sections. $N$ is first converted to gray-scale, then both $B$ and $N$ are passed to a Gaussian filter ($B \times g(\mu, \sigma)$ and $N \times g(\mu, \sigma)$) with $\sigma = 1$, to normalize brain sections with respect to the background noise. The corresponding reference atlas ($R$) of Nissl section is also down sampled to the same ratio, $r$.

After preprocessing, $N$ is registered and transformed to the dimensions of $B$ and the final metric score (error of cost function) is calculated. The same procedure is repeated $n$ times and the final metric score is plotted against every registration step as shown in **supplementary figure 6** to find the minima of error, here $n$=20. Intermediate steps of registration for the first ten iterations in medial and lateral sagittal sections are shown in (**supplementary figure 4 (a)**) and (**supplementary figure 5 (a)**), respectively. The corresponding image, $N$, at the minima of cost function is used for the purpose of registration. The transform parameter map is applied to the corresponding reference atlas, $R$. **Supplementary figure 4 (f)** and **supplementary figure 5 (f)** show the reference atlas, $R$, overlaid on top of original brain section, $B$ after registration, whereas **supplementary figure 4(d)** and **supplementary figure 5(d)** show the overlaid reference atlas before registration.

**Neural dataset generation for training the Faster RCNN.** A large dataset of six different brains is collected from Allen Brain's OPR, in total 220 (P4 GAD1: 16+16, P4 VGAT: 18+18, P14 GAD1: 17+17, P14 VGAT: 19+19, P56 GAD1: 20+20 and P56 VGAT: 20+20) high resolution images that includes ISH and Expression sections. A pair of two brains from each post-natal development time points (P4, P14, P56) is selected. For each brain, both ISH and expression images are utilized. ISH images are used for the classification of neurons and the corresponding expression images are utilized for applying the color threshold filter in order to distinguish the neural signal from background noise. Initially, selected sagittal brain sections in high-resolution are segmented into smaller images by equally dividing a sagittal brain section into 100×100 smaller sections, whereas each smaller image contains ~20 neurons. These small image dataset is transferred to a specific directory which is imported in *SiGUI*. In total, 220 randomly selected sections are used for human annotation, half of them (110) are randomly used for training and the other half (110) for testing purpose. **Supplementary figure 7** shows the small brain sections, followed by their expression versions in **supplementary figure 8**. **Supplementary figure 9** shows the expression images overlaid on top of the ISH brain sections.

**Ground-truth labelling.** A Simple Graphical User Interface (*SiGUI)* software is developed in MATLAB to generate the ground-truth data. Ground truth labels were generated by three human experts who annotated the bounding boxes on top of the neurons as shown in **supplementary figure 10**. Users can freely scroll towards left and right in the directory to jump on different images and draw and remove bounding boxes at their ease, by selecting the draw/remove bounding box options from the right panel of the software. The expression versions of the same brain sections were used as a reference to differentiate the real neural signal with a background noise (**supplementary figure 8**).

*DeNeRD*'s **Faster RCNN architecture.** The architecture of our network is inspired by the Faster RCNN [8] as shown in **supplementary figure 11**. Designing of the network architecture, training and testing is performed in MATLAB. A four-step training procedure is applied that includes: (i) Training of the Region Proposal Network (RPN), (ii) Use the RPN from (i) to train a fast RCNN, (iii) Retraining RPN by sharing weights with fast RCNN of (ii). Finally, (iv) Retraining Fast RCNN using updated RPN. Epoch size of 500 is used with initial learning rate of $1\times10^{-5}$ and $1\times10^{-6}$ in stages (i-ii) and (iii-iv), respectively. Furthermore, box pyramid scale value of 1.2 is set with positive and negative overlapping ranges of [0.5 1] and [0 0.5]. Number of box pyramidal levels are set to 15 with the minimum object size of [2, 2]. For all of the convolutional stages, padding size of [1, 1, 1, 1] is selected with [1, 1] as the size of strides whereas a grid size of [2, 2] is set in the RoI pooling layer. Training is performed by using NVIDIA Quadro P4000 GPU. Our network is trained by minimizing the multi-task loss, which corresponds to each labeled Region of Interest (RoI) (i.e. neuronal body), through stochastic gradient descent algorithm [20]. **Supplementary figure 12** shows the network learning the features of neurons (weight vector) after training. Similar to [8], loss of our network is described as following:

$$L = L_{cls} + L_{reg}$$

Here, $L_{cls}$ = classification loss of neuron; calculated as a log loss for neuron *vs.* non-neuron labelled classes, and $L_{reg}$ = regression loss of bounding box, where $L_{cls}$ and $L_{reg}$ are defined by following:

$$L(p_i, q_i) = \frac{1}{n_{cls}} \sum_i L_{cls}(p_i, p_i^*) + \lambda \frac{1}{n_{reg}} \sum_i L_{reg}(q_i, q_i^*)$$

$p_i$ = proposed $i^{th}$ RoI's (or anchor) probability
$p_i^* = \begin{cases} 1, & p_i \geq 0.7 \\ 0, & p_i \leq 0.3 \end{cases}$
$q_i$ = bounding box coordinates of predicted RoI/anchor
$q_i^*$ = bounding box coordinates of a positive anchor
$n_{cls}$ = normalization parameter of classification loss, $L_{cls}$
$n_{reg}$ = normalization parameter of regression loss, $L_{reg}$
$\lambda$ = weight parameter

**Neural detector pipeline.** Each brain section is serially passed to the neural detector. After some pre-processing, an input brain section ($B$) of size ($p \times q$) pixels is divided into smaller brain sections ($b_n$) by dividing $B$ into ($s \times t$) equal sections of size, where $s=t=100$. Zero padding is performed on each section in $b_n$, hence the dimensions of a single image, $b_i$, is increased to ($s+1 \times t+1$). Now each $b_i$ ($i = 1 \rightarrow n$) is passed through the Faster R-CNN (**supplementary figure 11(a)**) which detects the neurons in each small section and returns the corresponding bounding boxes against each neural location. Finally, a binary image ($a$) of the same dimension ($s+1 \times t+1$) as $b_i$ is generated, whereas $a \in \{0,1\}$. The center of each bounding box location in a given $b_i$ is set to 1 in $a$ and 0 otherwise. After removing the zero padding, $a$ is stored with the corresponding iteration $i$ and the same process is repeated until the $n$th brain section. Once all the smaller sections ($b_{1 \rightarrow n}$) are passed through the detector and the corresponding binary sections ($a_{1 \rightarrow n}$) are generated, these binary sections are concatenated as following to generate a single binary image ($A$) with neural locations as 1's and background pixels as 0's. The dimension of $A$ will be the same as the input brain section, $p \times q$.

$$A_{(p \times q)} = a_1 | a_2 | a_3 | a_4 \ldots | a_n$$

**Neural density in developing brains.** Registered reference atlases ($R_i$) of brains are collected from brain registrator, and binary neural images ($A_i$) are imported from neural detector pipeline. Each brain region ($R_i$) has a unique RGB color code, brain regions ($R_N$) are filtered by their respective RGB codes and temporarily stored in the form of binary images where active $R_i$ pixels are 1s, and 0s otherwise:

$$R_i(x) = \begin{cases} 1, & x = [r_i, g_i, b_i] \\ 0, & x \neq [r_i, g_i, b_i] \end{cases}$$

Afterwards, neural density ($D_i$) of a particular brain region ($R_i$) is calculated by taking its dot product with the binary neural image of a given brain section ($A_i$) and normalize it by dividing with the area of the brain region, $|R_i|$.

$$D_i = \frac{R_i \cdot A_i}{|R_i|}$$

The same process is repeated for rest of the brain regions ($R_{N-1}$) on a given $B_i$. Once, the neural density is measured for all the regions in a single brain section ($A_i$), the next brain section in the pipeline is analyzed and so on until $A_N$ for a particular age, where $N$ denotes the total number of brain regions. Similarly, neural density of the same brain regions at various ages (P4, P14 and P56) is computed. **Supplementary figures 13-18** show the detected neurons after passing through neural detector and brain registrator pipelines in complete brain datasets across development.

**Statistics.** Wilcoxon rank-sum test [21] is used to measure the statistical difference between the neural density of entire brain populations (**figure 3**). It is a non-parametric test which is used to determine if the two population samples have the same distribution.

**Software availability.**
The code and datasets generated during and/or analyzed during the current study are available from the corresponding author on reasonable request. *DeNeRD* pipeline will be available online on bitbucket (https://bitbucket.org/theolab/) for the neuroscience community.

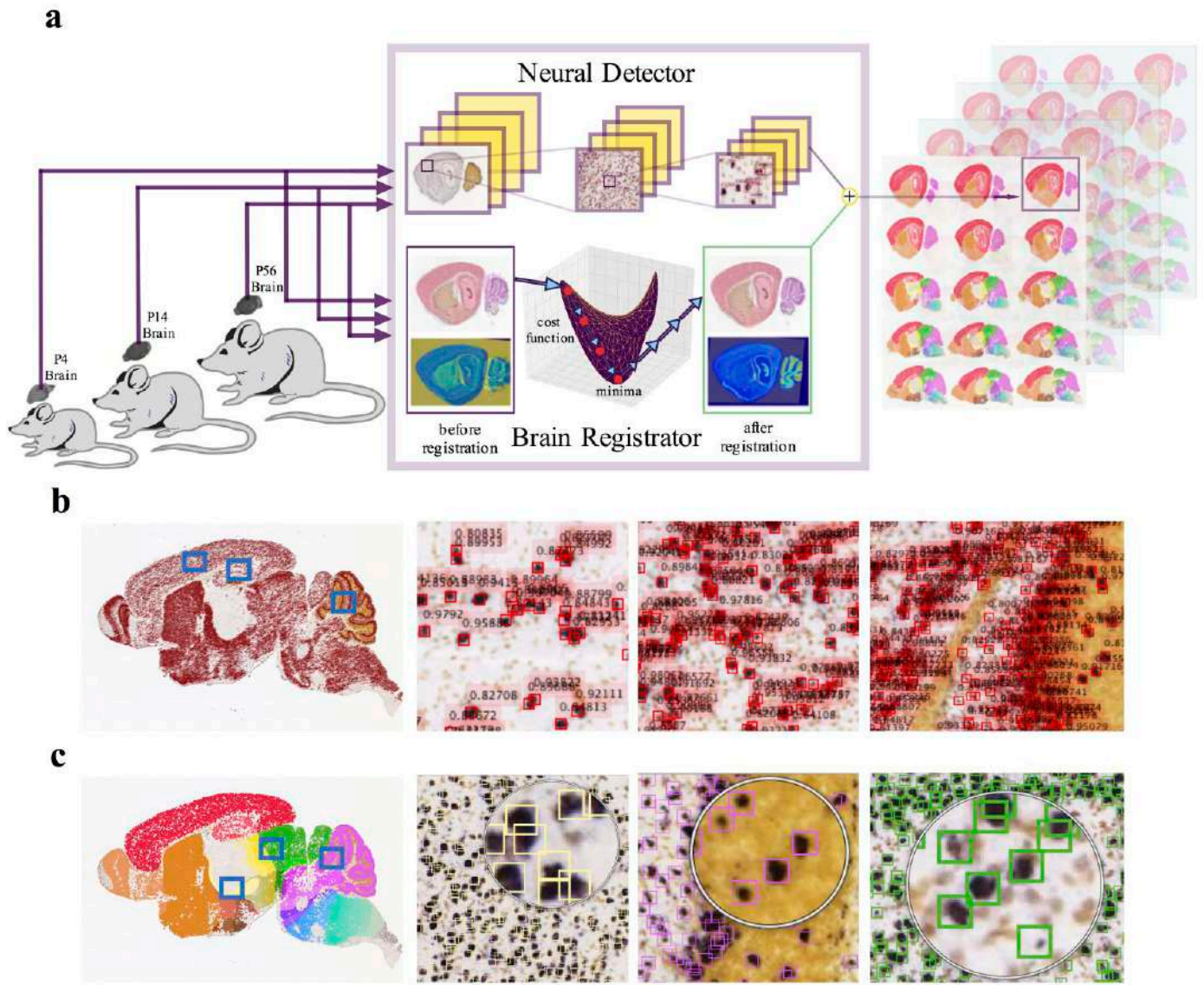

**Fig. 1 | Block diagram architecture and functionality of deep learning-based system, *DeNeRD*.**
**a,** Input brain images of two different markers (GAD1 and VGAT) and at three different postnatal time points P4, P14, P56 are fed to the system. Brain sections are processed by the *Neuron Detector* (top) unit where each brain section is serially passed through Faster Regions with Convolutional Neural Network (R-CNN) architecture which detects and labels all the neurons present in the brain section. The same section is also passed through the *Brain Registrator* (bottom) unit which registers the brain section with the Allen developing mouse brain reference atlas. Registration is performed through the optimization of an affine transformation algorithm. Output of *Neuron Detector* and *Brain Registrator* is combined to label all the detected neurons based on the color-coded mouse brain atlas regions. Output on the right side shows the clusters of neurons in different brain regions in both lateral and medial brain sections of P14 GAD1 mouse.
**b,** Output of the *Neuron Detector* on a medial section (sagittal plane) of a P14 mouse brain is shown, neurons are detected with their classification scores over the bounding boxes. Performance of the *Neuron Detector* at three zoomed-in examples of low to high dense neural regions are shown at the right side.
**c,** The same brain section in fig. 1 (b) is passed through the *Brain Registrator* module, bounding boxes of detected neurons are now labelled with their color coded brain regions. Performance of *Neuron Detector* is also shown on the right side at three randomly selected zoomed-in brain regions (pre-thalamus, hindbrain and midbrain).

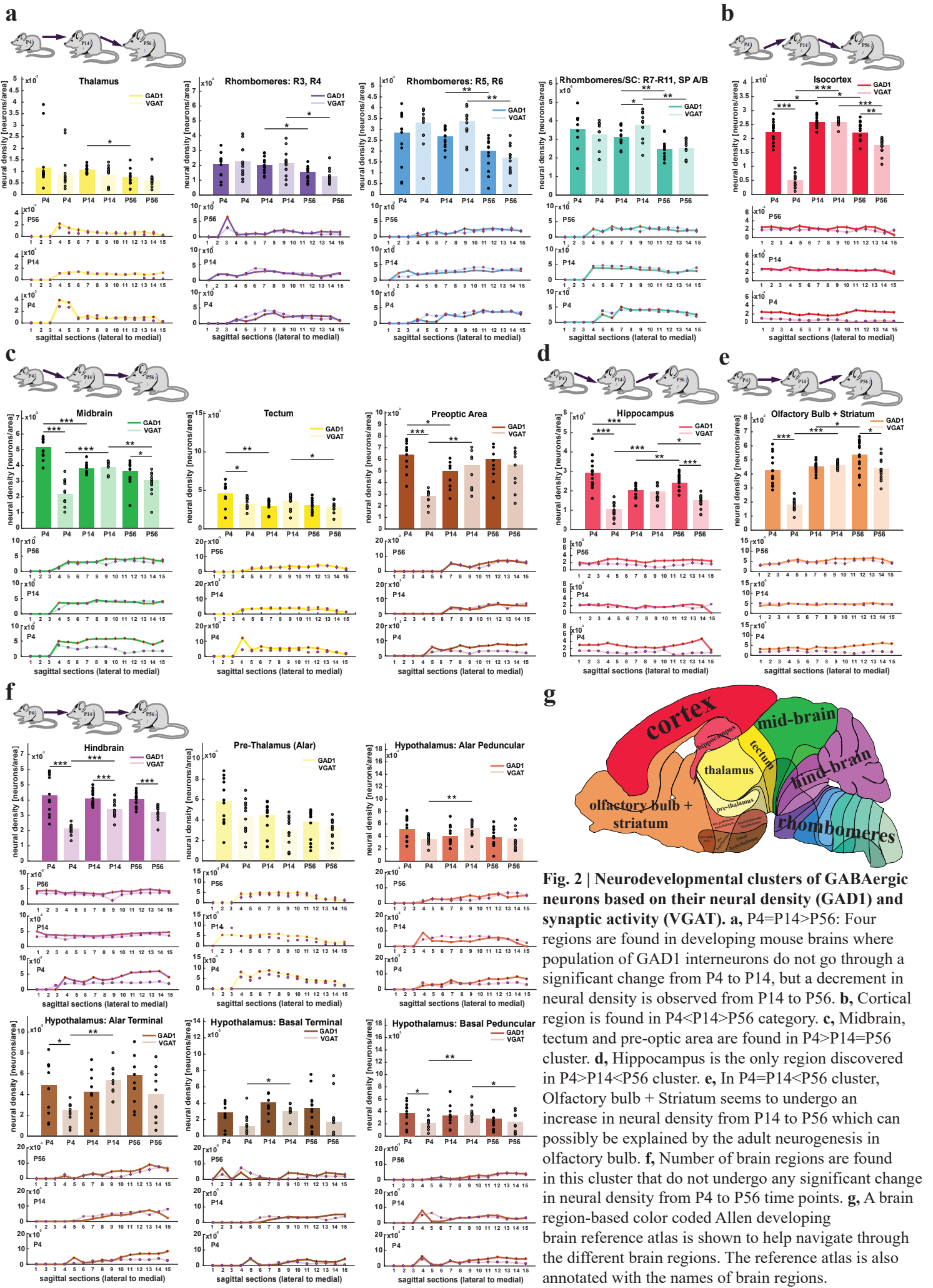

Fig. 2 | Neurodevelopmental clusters of GABAergic neurons based on their neural density (GAD1) and synaptic activity (VGAT). a, P4=P14>P56: Four regions are found in developing mouse brains where population of GAD1 interneurons do not go through a significant change from P4 to P14, but a decrement in neural density is observed from P14 to P56. b, Cortical region is found in P4<P14>P56 category. c, Midbrain, tectum and pre-optic area are found in P4>P14=P56 cluster. d, Hippocampus is the only region discovered in P4>P14<P56 cluster. e, In P4=P14<P56 cluster, Olfactory bulb + Striatum seems to undergo an increase in neural density from P14 to P56 which can possibly be explained by the adult neurogenesis in olfactory bulb. f, Number of brain regions are found in this cluster that do not undergo any significant change in neural density from P4 to P56 time points. g, A brain region-based color coded Allen developing brain reference atlas is shown to help navigate through the different brain regions. The reference atlas is also annotated with the names of brain regions.

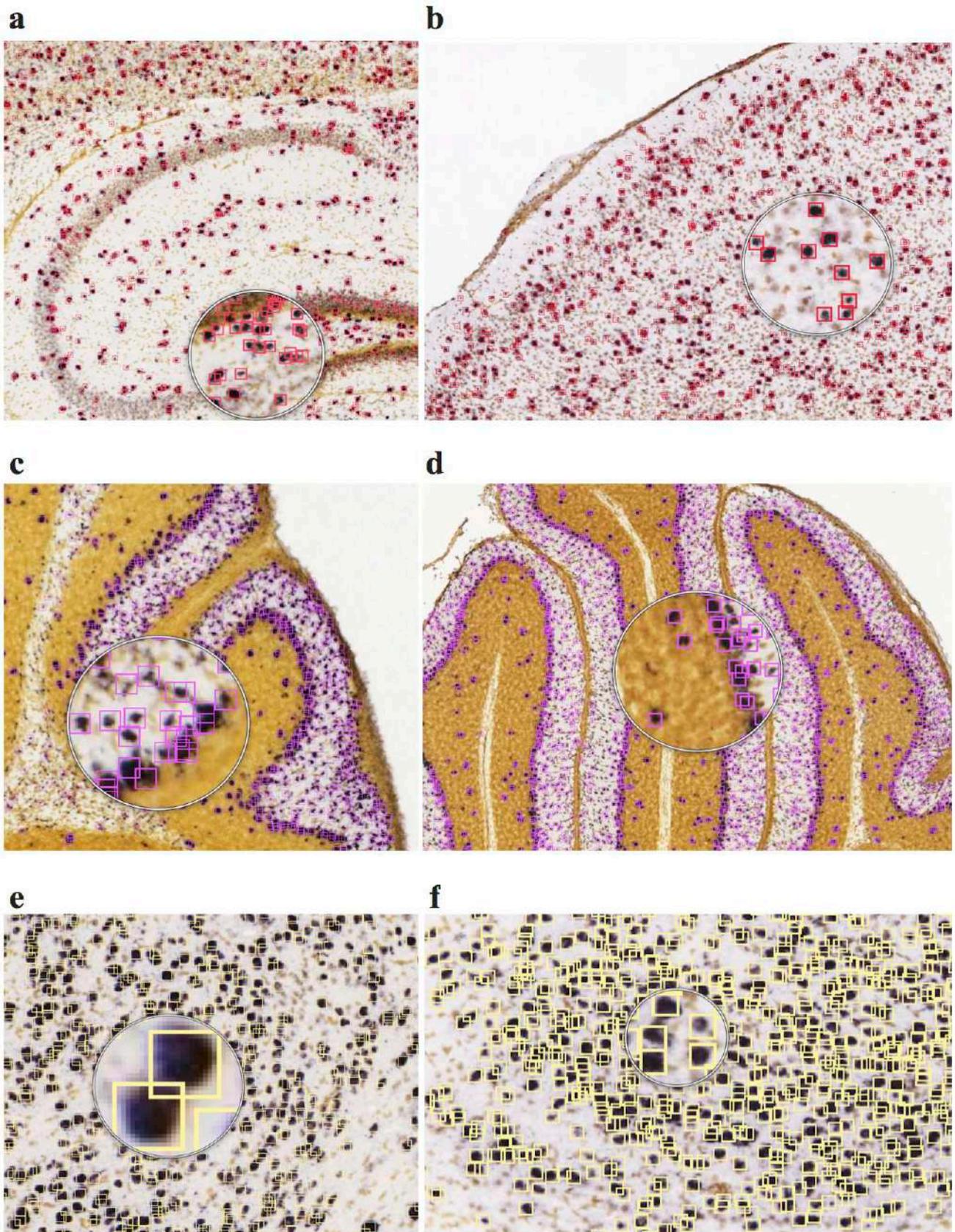

**Supplementary Figure 1. Performance of DeNeRD on brain sections of various ages.**
**a,** Network is detecting neurons in a section captured from P4 VGAT brain. RCNN can localise neurons (zoomed) in a variety of background conditions in hippocampus region. **b,** Neurons are detected in cortex of P56 GAD1 mouse brain. Network is able to differentiate neural signal from background signals (zoomed). **c-d,** Performance of network in detecting neurons with a high contrast in background and neural structure is shown in P4 and P14 mouse brain tissues (zoomed) at hindbrain region. **e-f,** Network is able to detect neurons in a dense region of pre-thalamus at P56 and P14 mice brain tissues (zoomed).

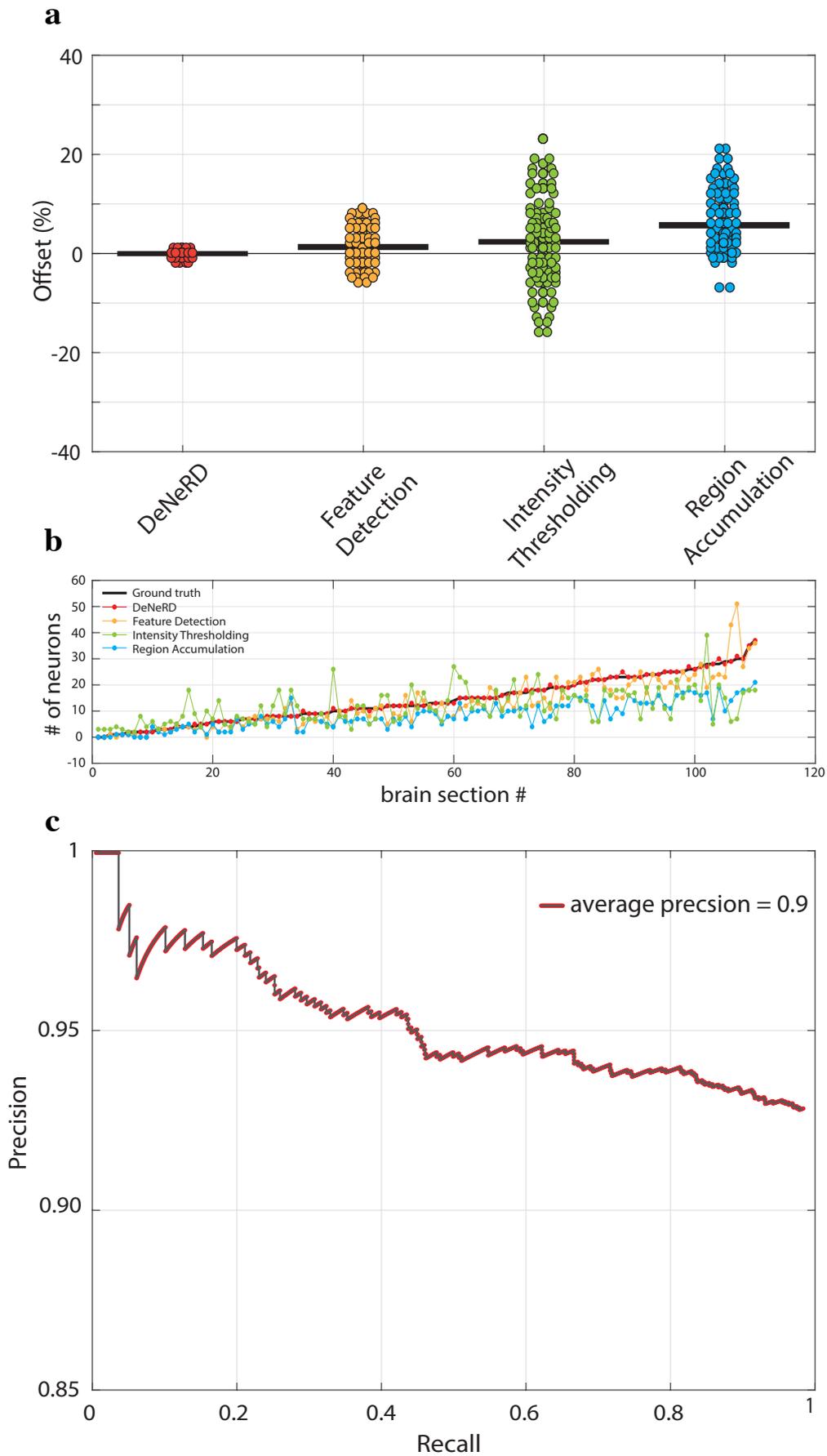

**Supplementary Figure 2. Performance of DeNeRD in comparison with other methods.**
a, Performance comparison of DeNeRD with commonly used cell counting methods. DeNeRD outperforms the existing methods by giving a minimum offset on test samples. b, Performance comparison with other methods on testing data with increasing neural density. c, Precision-Recall curve of our method on testing dataset. Deep Neural Network achieves a high score in detecting neurons with an average precision of 0.9.

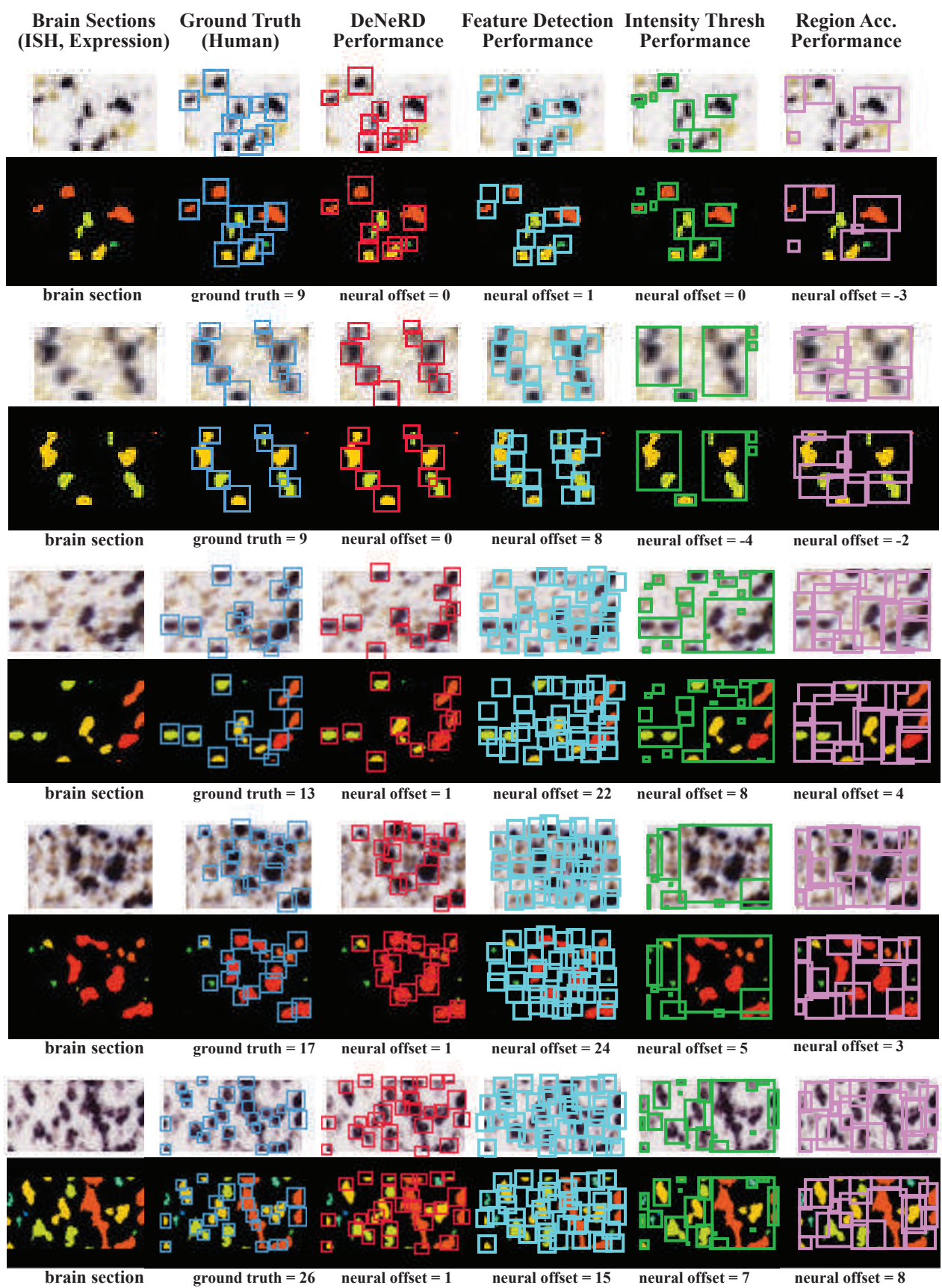

**Supplementary Figure 3. Comparision of *DeNeRD* with commonly used cell detection techniques.** Left column shows five brain sections (In-Situ Hybridized and Expression) as examples wheres the second column shows the ground-truth (human annotated) bounding boxes. Third column shows the performance of DeNeRD on the given sections, followed by the performances of Feature Detection, Intensity Thresholding and Region Accumulation-based (watershed) algorithms. DeNeRD outperforms these methods in almost all the examples. For <10 # of neurons in a given section (row # 1-4) the other three methods show relatively low neural offset score but in examples containing >10 # of neurons, the neural offset score of the other three methods gets higher.

a

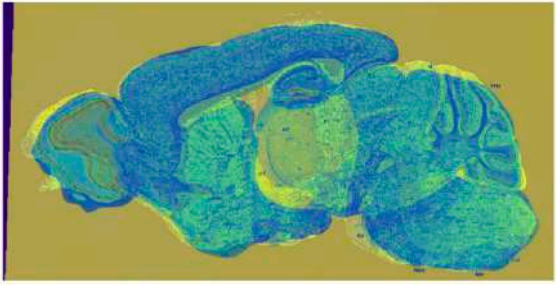 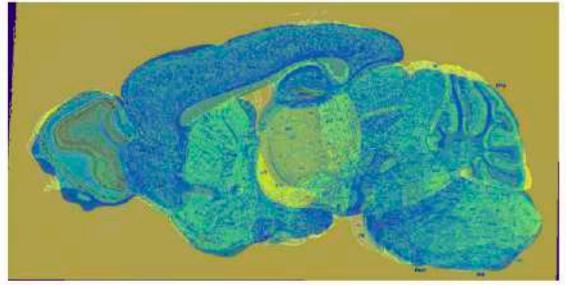

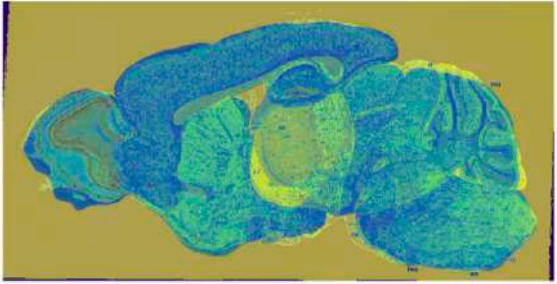 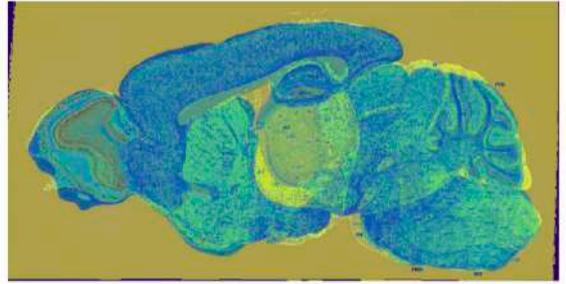

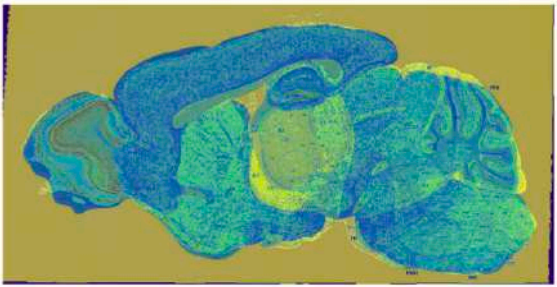 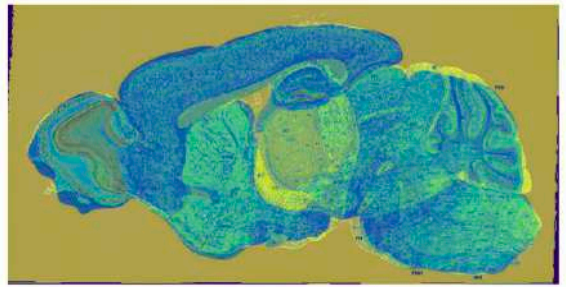

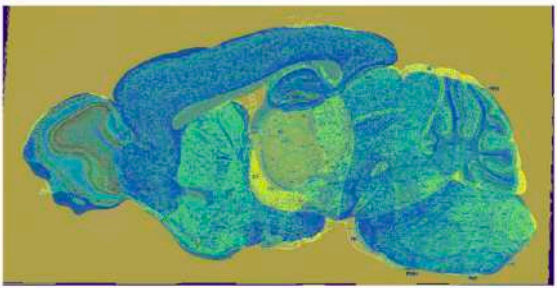 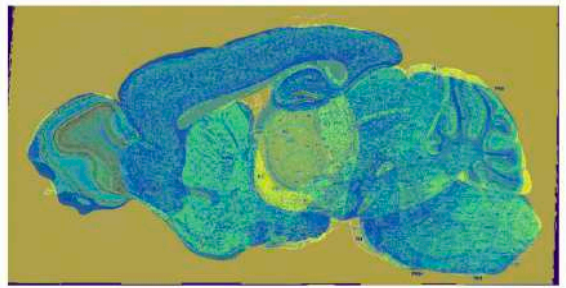

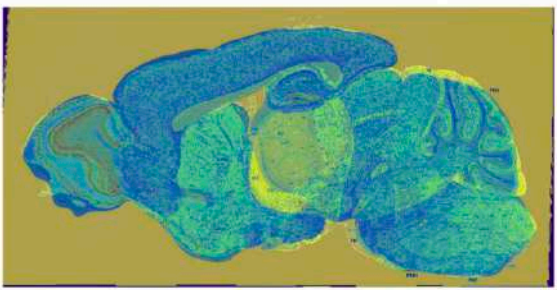 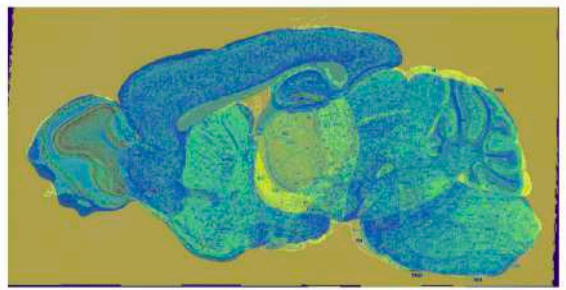

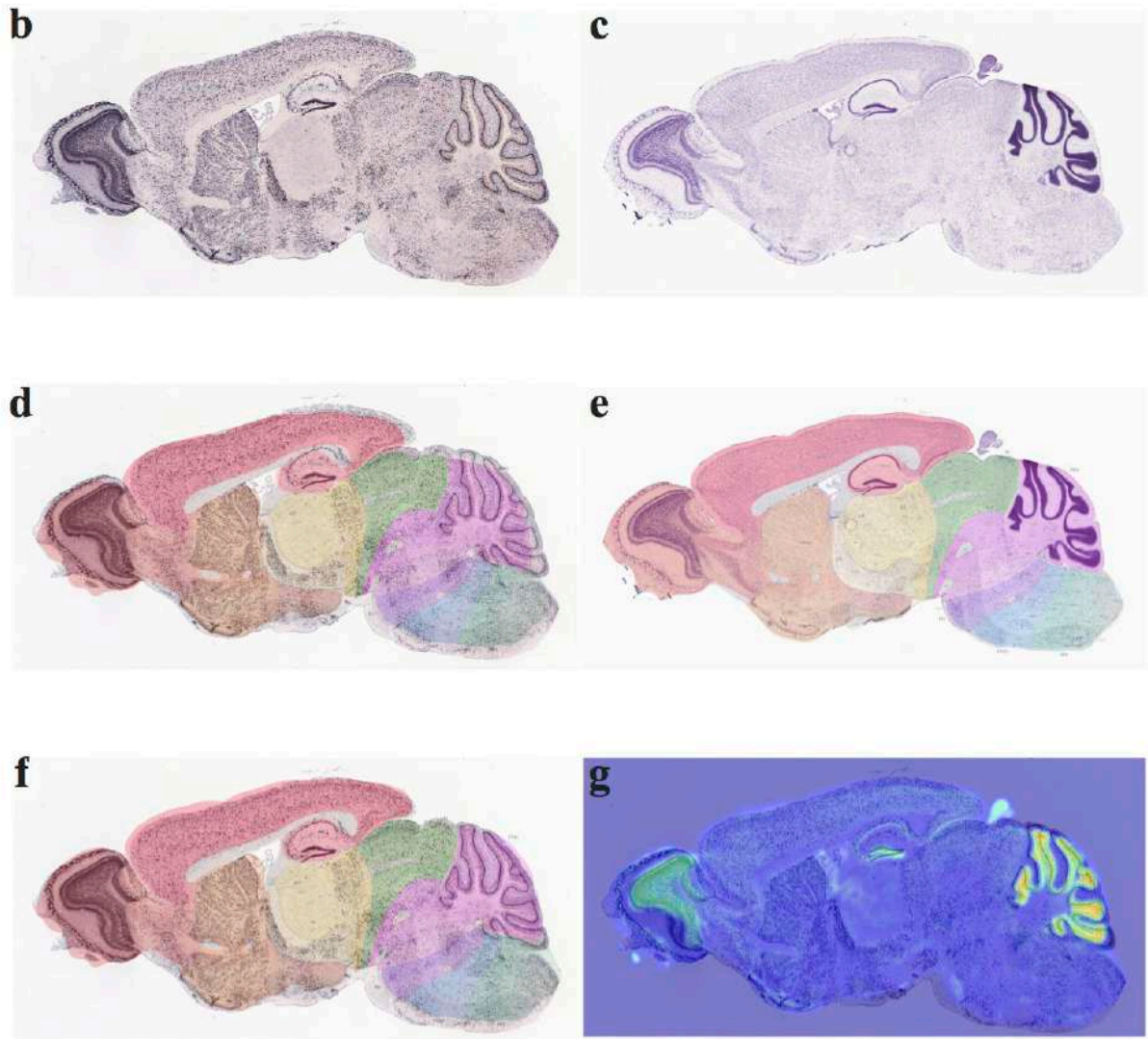

**Supplementary Figure 4. Registration of a medial brain section (sagittal) of an adult brain with Allen brain developing reference atlas.**

**a,** 10 iterations of registration by optimisation of affine transformation algorithm in ITK. **b,** Sample lateral (sagittal) section of VGAT marker from an adult mouse brain. **c,** Nissl brain section against the corresponding brain section in (b). **d,** Overlaid reference atlas on top of brain section in (b) before registration. **e,** Overlaid reference atlas on top of a Nissl brain section. Atlas is pre-registered with the Nissl section. **f,** Overlaid transformed reference atlas on top of the VGAT brain section in (b) after registration. **g,** Overlaid VGAT brain section with transformed Nissl brain section after registration.

a

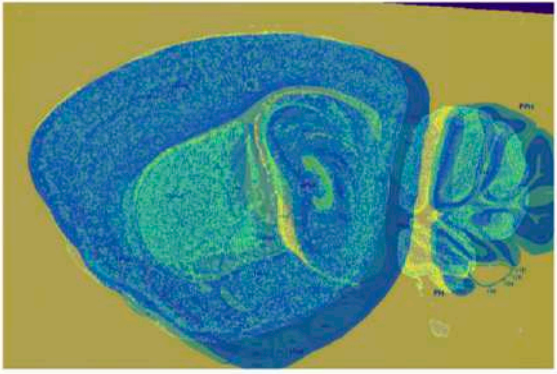 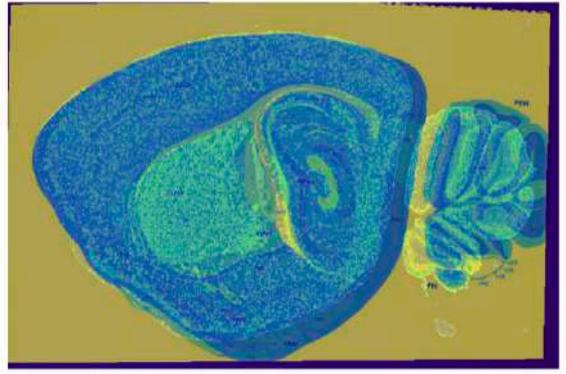

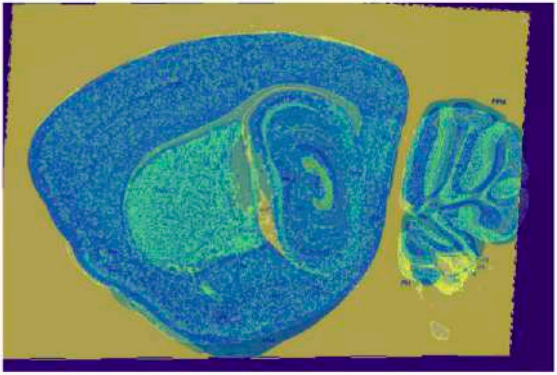 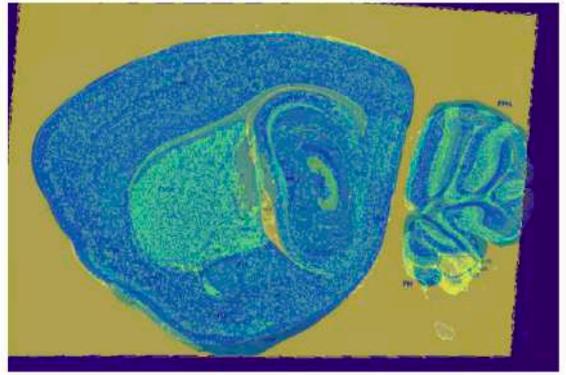

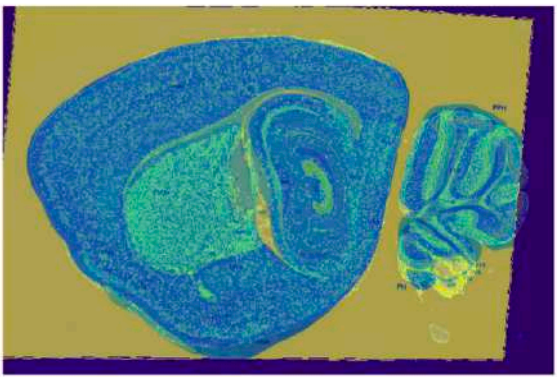 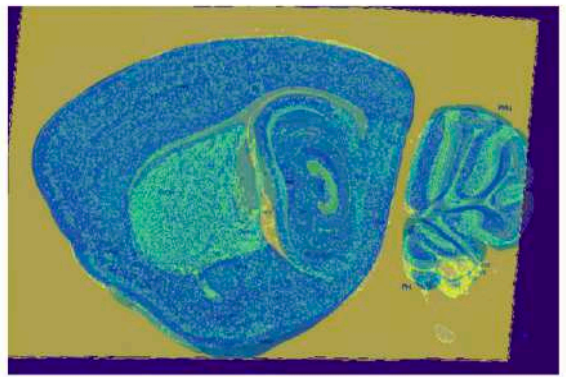

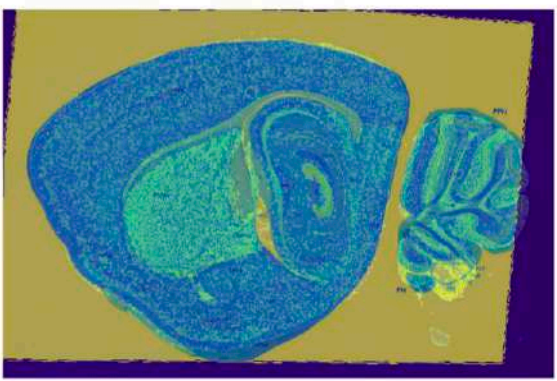 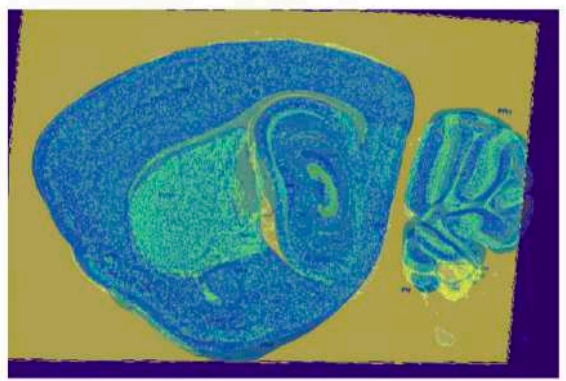

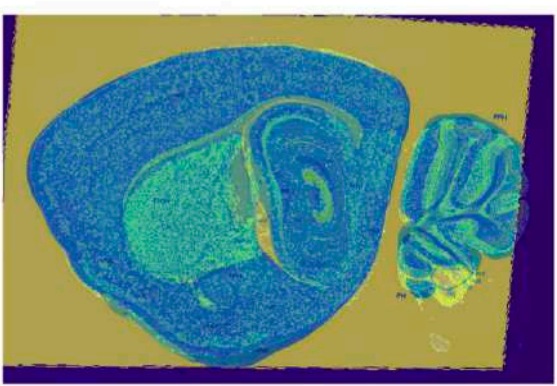 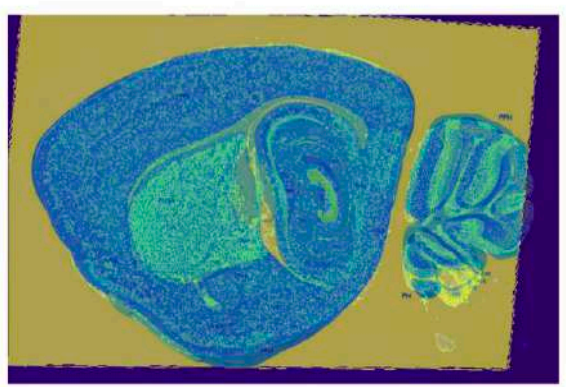

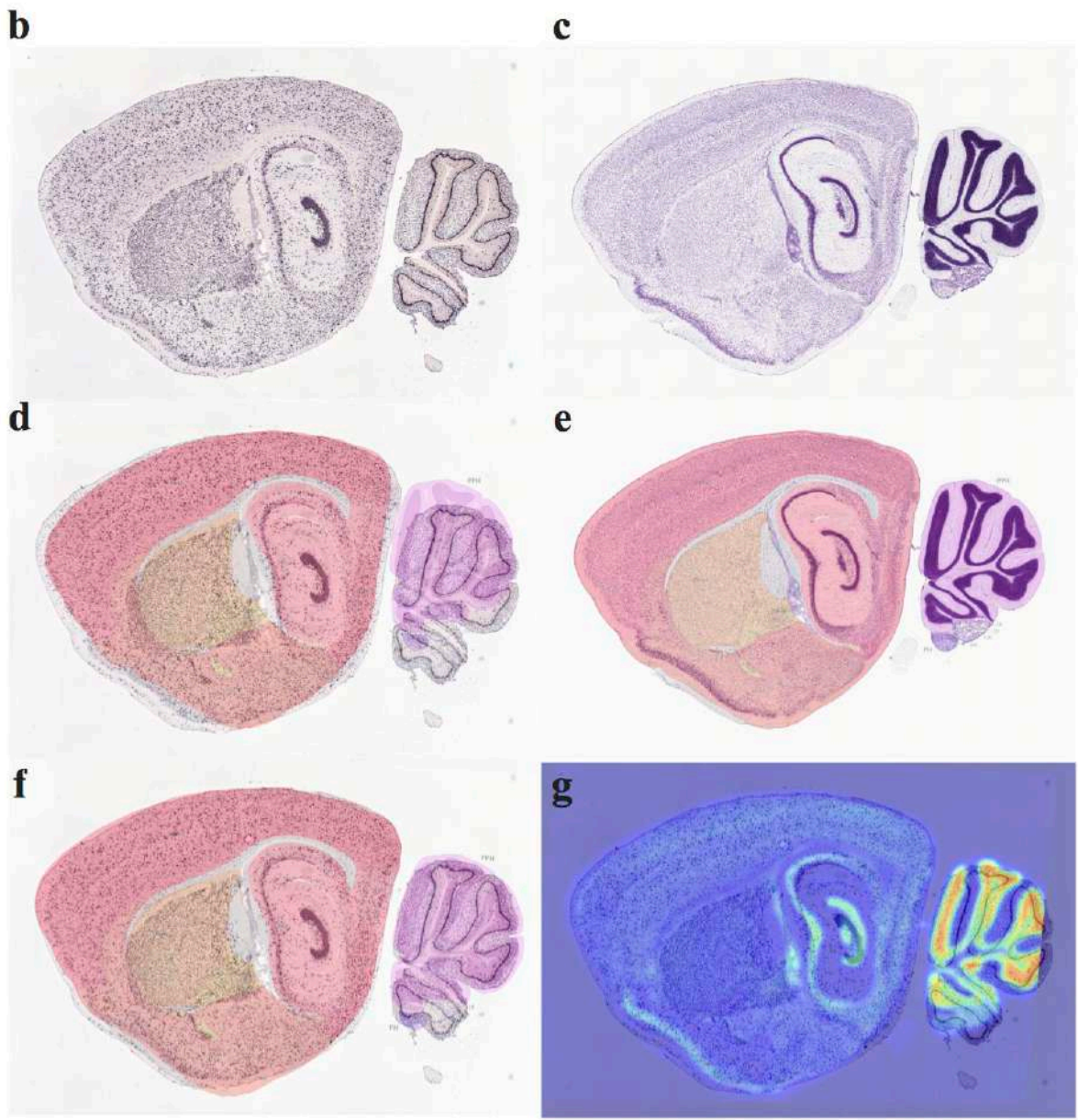

**Supplementary Figure 5. Registration of a lateral brain section (sagittal) of an adult brain with Allen brain developing reference atlas.**

**a,** 10 iterations of registration by optimisation of affine transformation algorithm in ITK. **b,** Sample lateral (sagittal) section of VGAT marker from an adult mouse brain. **c,** Nissl brain section against the corresponding brain section in (b). **d,** Overlaid reference atlas on top of brain section in (b) before registration. **e,** Overlaid reference atlas on top of a Nissl brain section. Atlas is pre-registered with the Nissl section. **f,** Overlaid transformed reference atlas on top of the VGAT brain section in (b) after registration. **g,** Overlaid VGAT brain section with transformed Nissl brain section after registration.

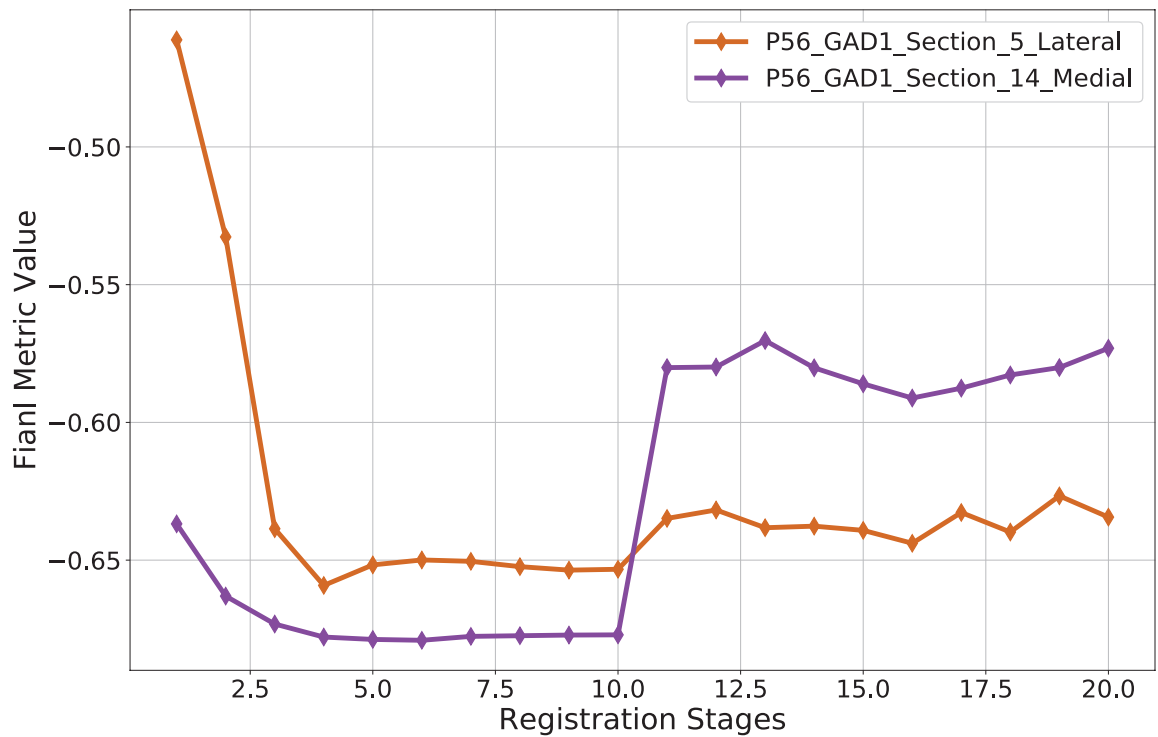

**Supplementary Figure 6. Error minimisation for Brain Registrator.**
A recurrent strategy is applied to optimise the brain registrator pipeline in order to register a given brain section with the corresponding reference atlas. A general sketch is previously shown in the brain registrator module of figure 1. The graph here shows the same procedure applied on two different brain sections: 1) lateral brain section of the adult mouse brain (brown) and 2) medial brain section of an adult mouse brain (purple). The final metric value (error) usually reaches the minimum in the first ten recurrences, and appears to be increasing in the later stages. The registered reference atlas having the minimum metric value is used for registration against the corresponding brain section.

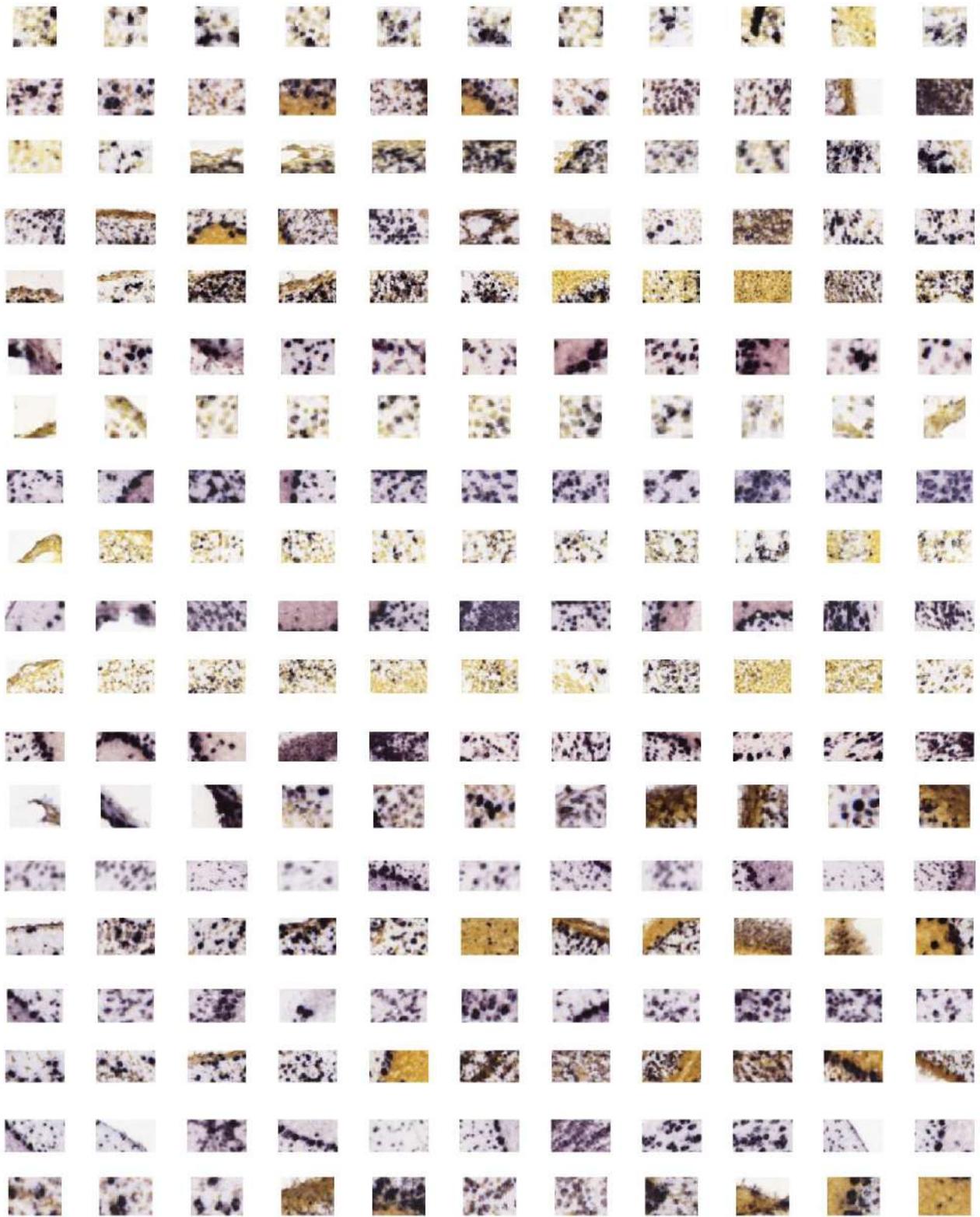

**Supplementary Figure 7. Brain sections to generate the ground-truth data.**
220 brain sections are randomly chosen from six different brains at three developing ages (P4, P14 and P56) of GAD1 and VGAT markers. These brain sections contain samples from various brain regions e.g. cortex, hippocampus, hindbrain, midbrain, etc.

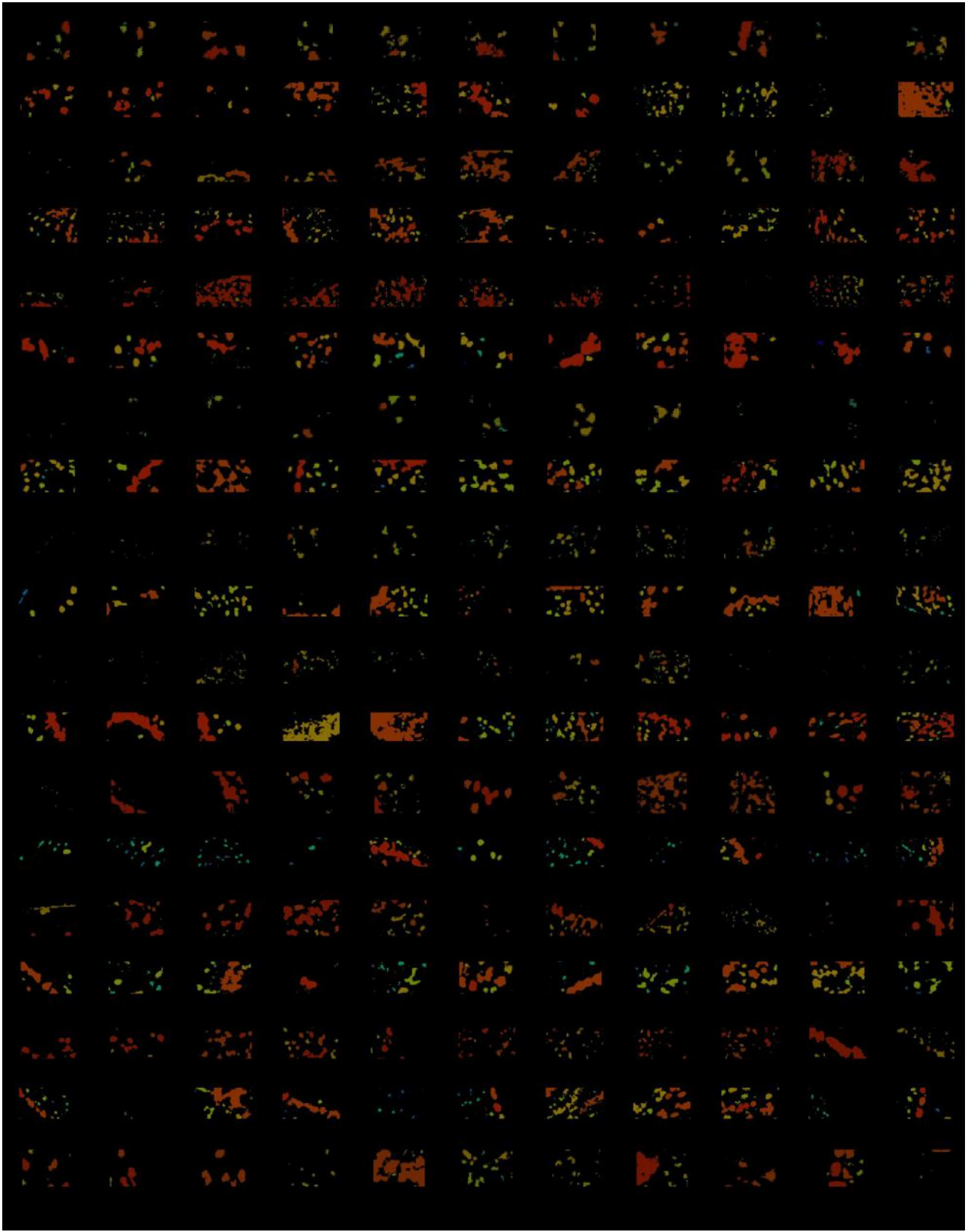

**Supplementary Figure 8. Brain sections to generate the ground-truth data (expression).** 220 samples as expression against the same brain sections in Supplementary figure 7 are shown. The neural expression is shown as a heat map where red is the strongest signal and blue is the weakest. These expression images are utilised to filter the real neurons from the background. Furthermore, human annotation is performed by looking at these expression samples as guidance to differentiate between the real neural signal with background structures.

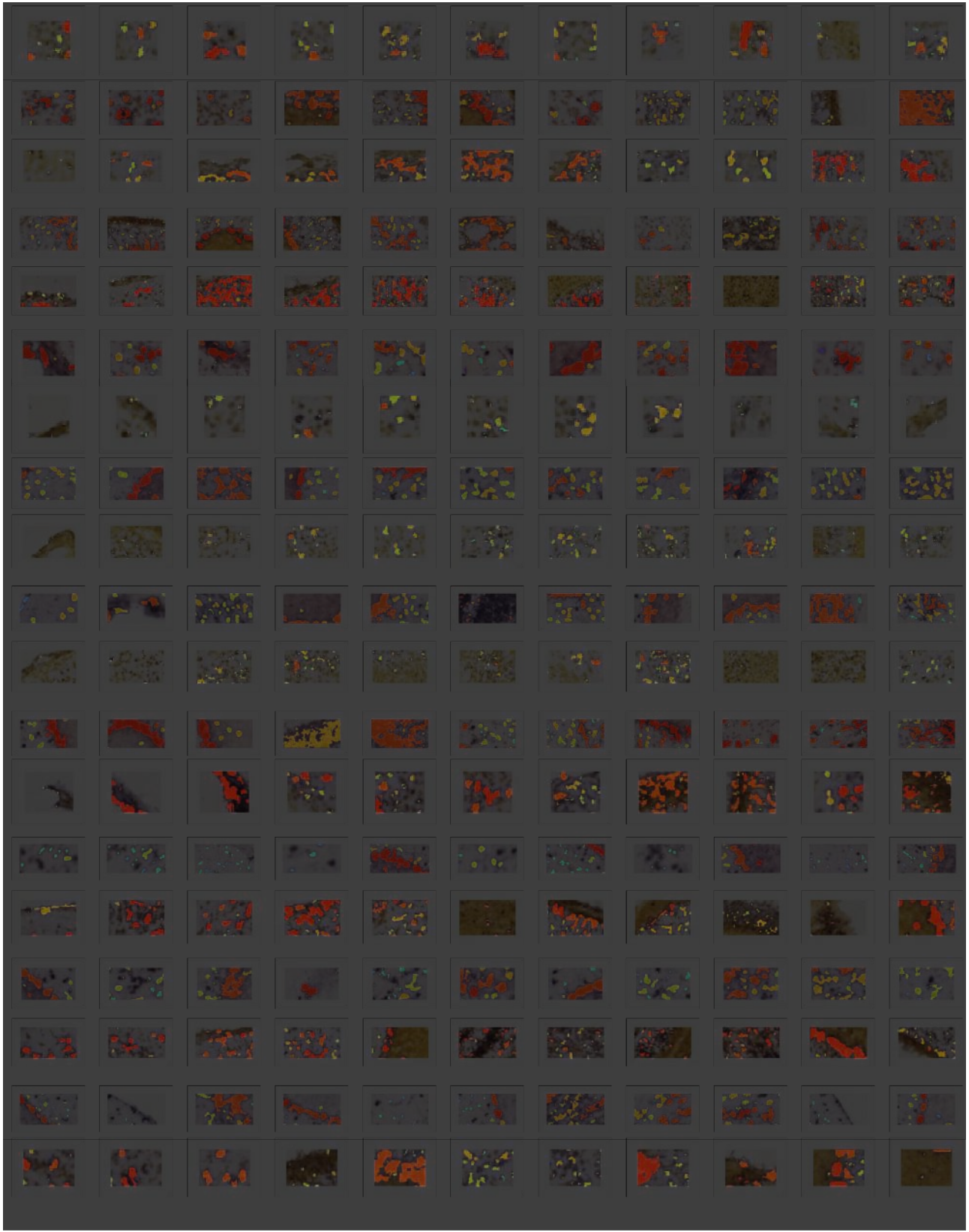

**Supplementary Figure 9. Overlay of expression images on top of ISH brain sections.** Images utilised for ground-truth training data of Faster RCNN are overlaid on top of their expression images in order to determine the real neural signal vs background. The same images are shown in supplementary figure 7 and 8.

**Supplementary Figure 10. Simple Graphical User Interface (SiGUI).**

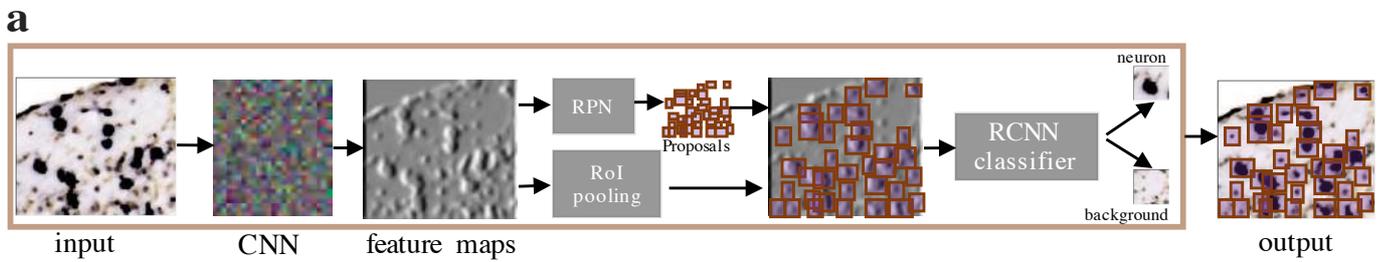

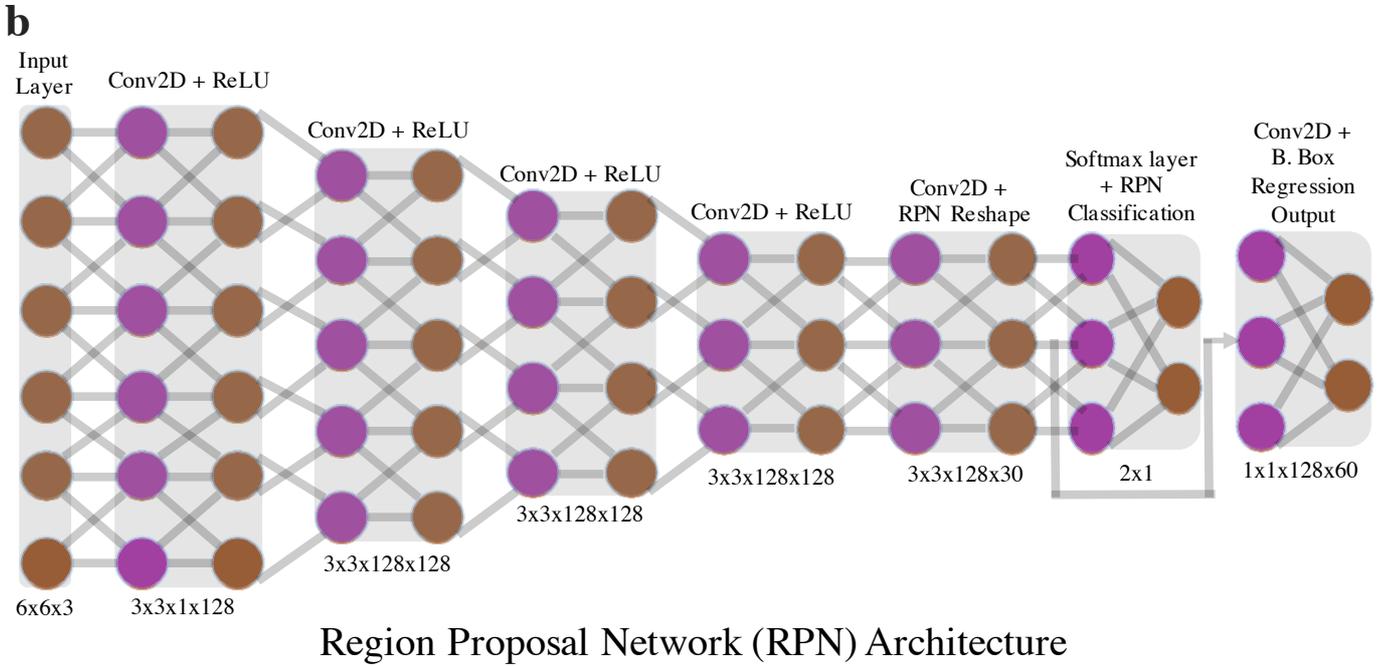

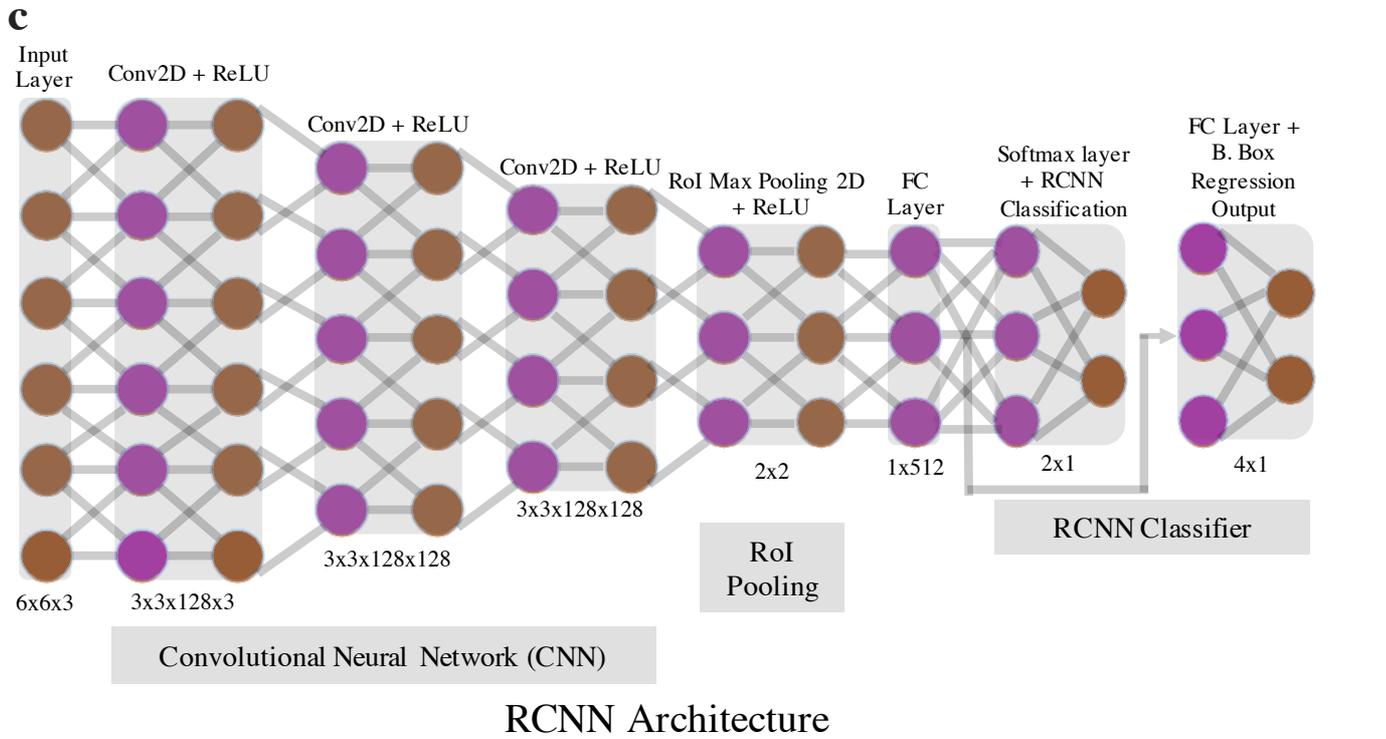

**Supplementary Figure 11. DeNeRD Architecture.**
a, Block diagram of Faster R-CNN scheme in DeNeRD. A small brain section is passed as input and it goes through various convolutional stages, RPN, RoI Pooling, and RCNN classifier. b, RPN Architecture with convolutional, ReLU and pooling layers. c, RCNN architecture with convolutional, ReLU, pooling and fully connected layers.

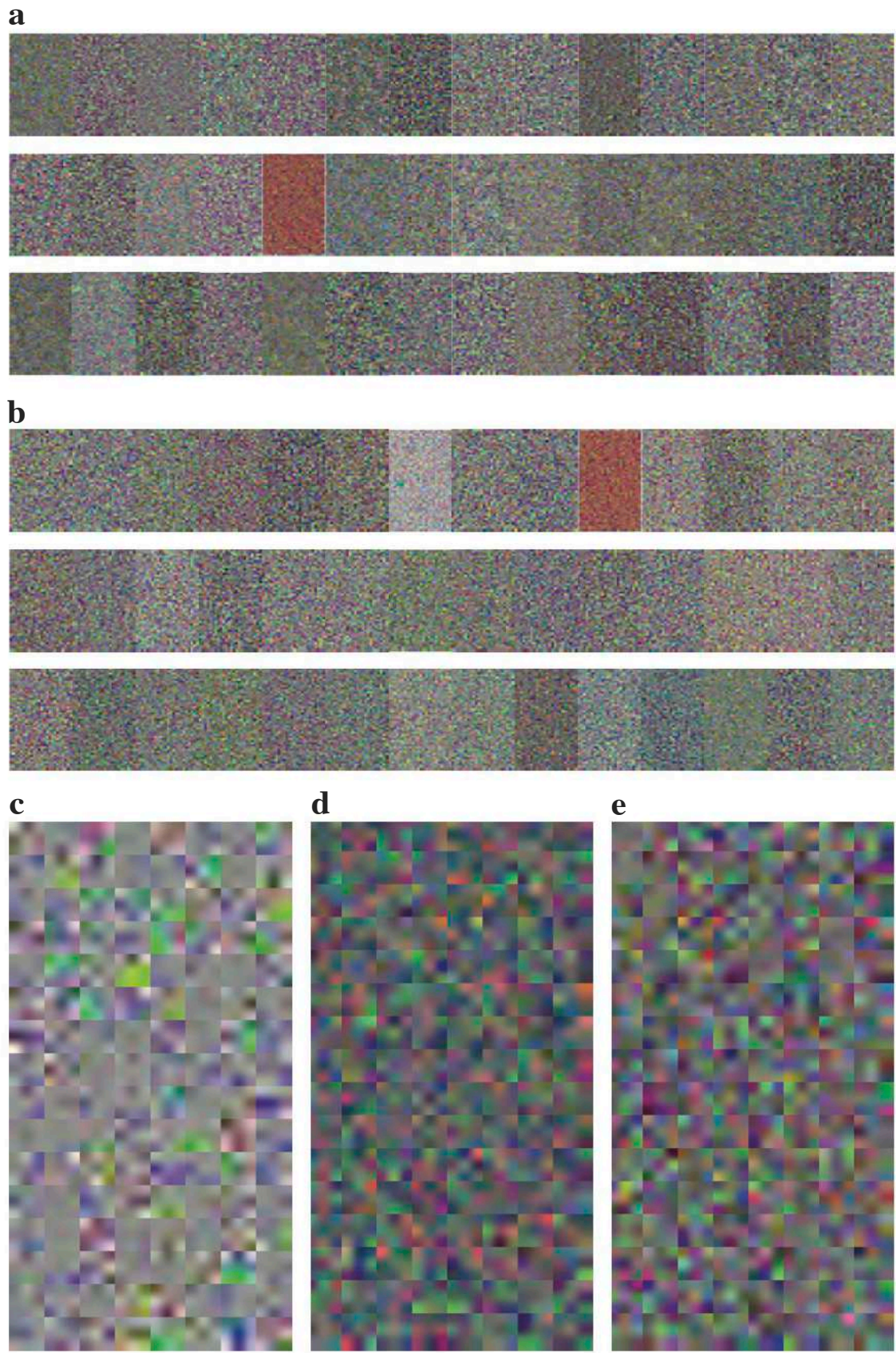

**Supplementary Figure 12. Weight vectors of convolutional layers in DeNeRD.**
a-c, Weight vectors of convolutional layers (conv4, conv6 and conv2 layers) in Faster RCNN architecture. d-e, A subset of weight vectors (highlighted in red of conv4 and conv6) are shown. Network seems to learn the structure of neurons in the process of training.

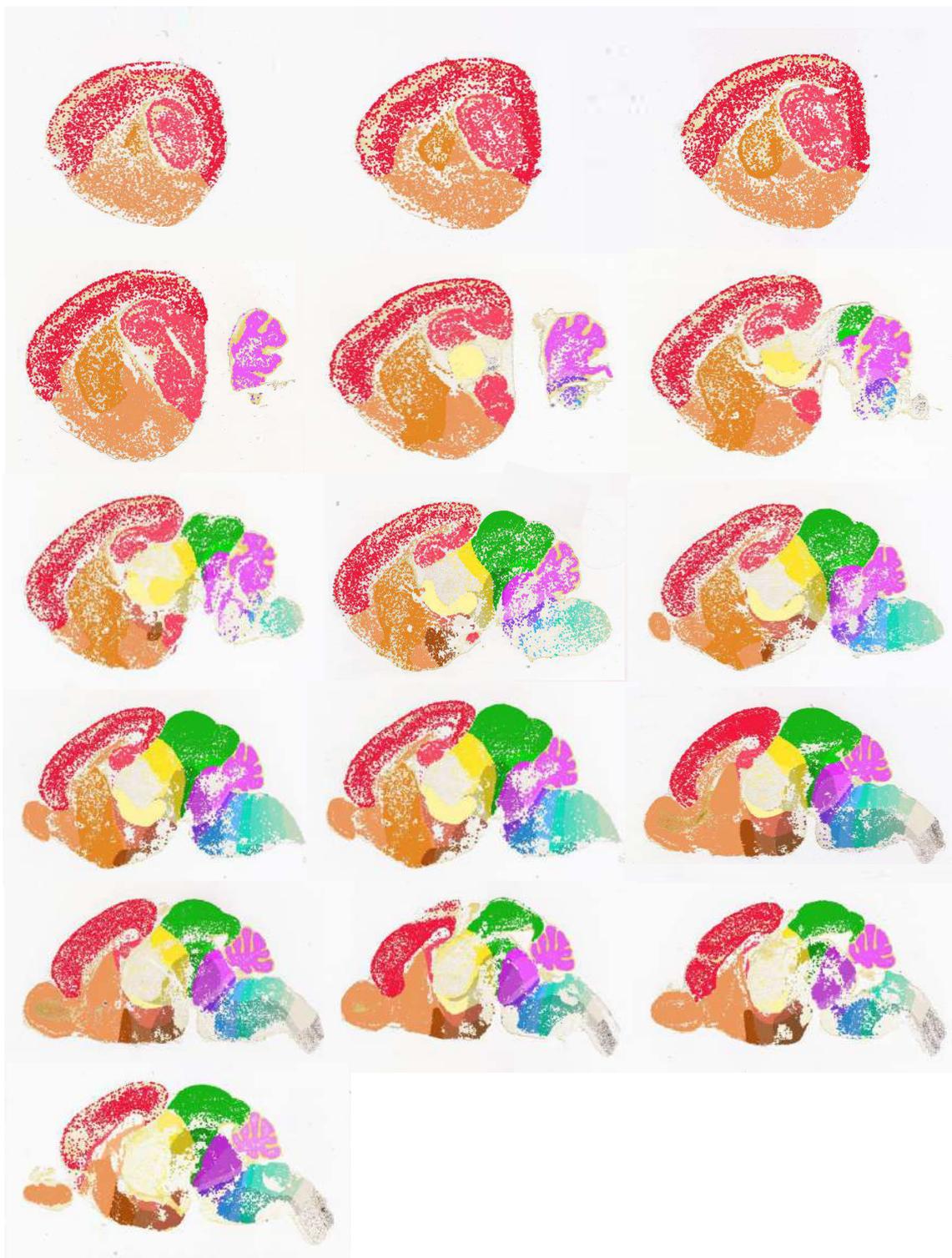

**Supplementary Figure 13. P4 GAD1 brain sections after passing through DeNeRD.**

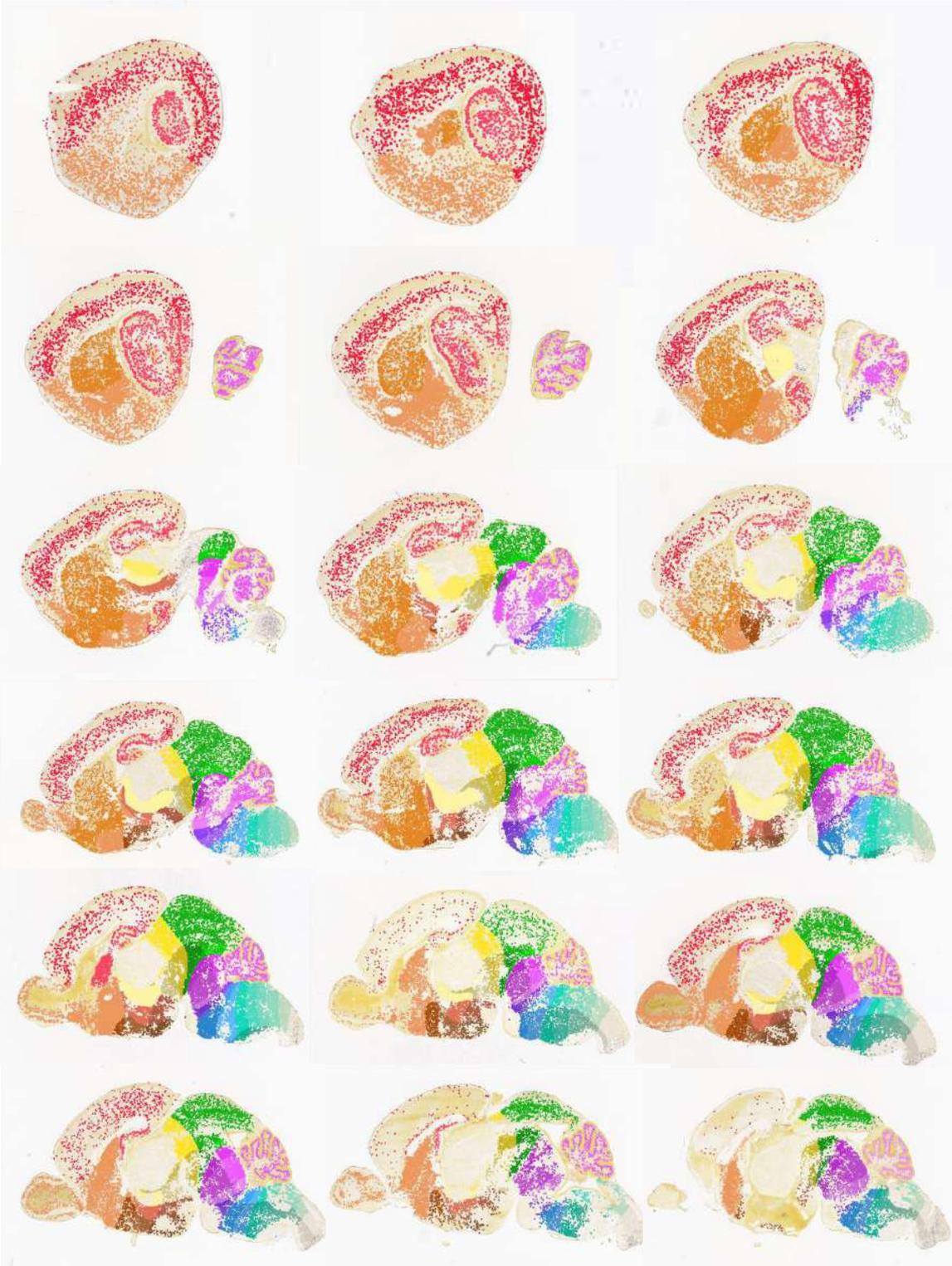

**Supplementary Figure 14. P4 VGAT brain sections after passing through DeNeRD.**

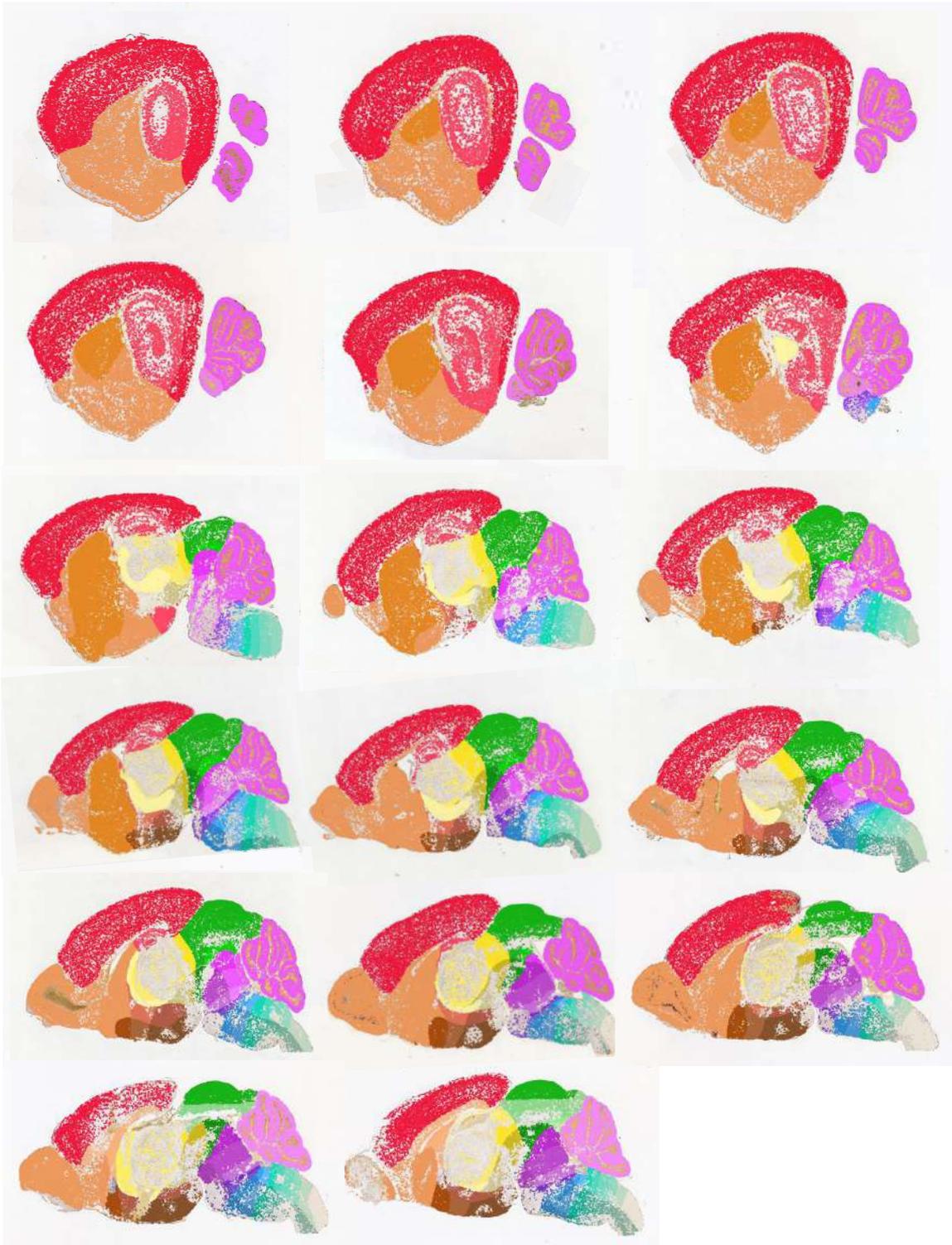

**Supplementary Figure 15. P14 GAD1 brain sections after passing through DeNeRD.**

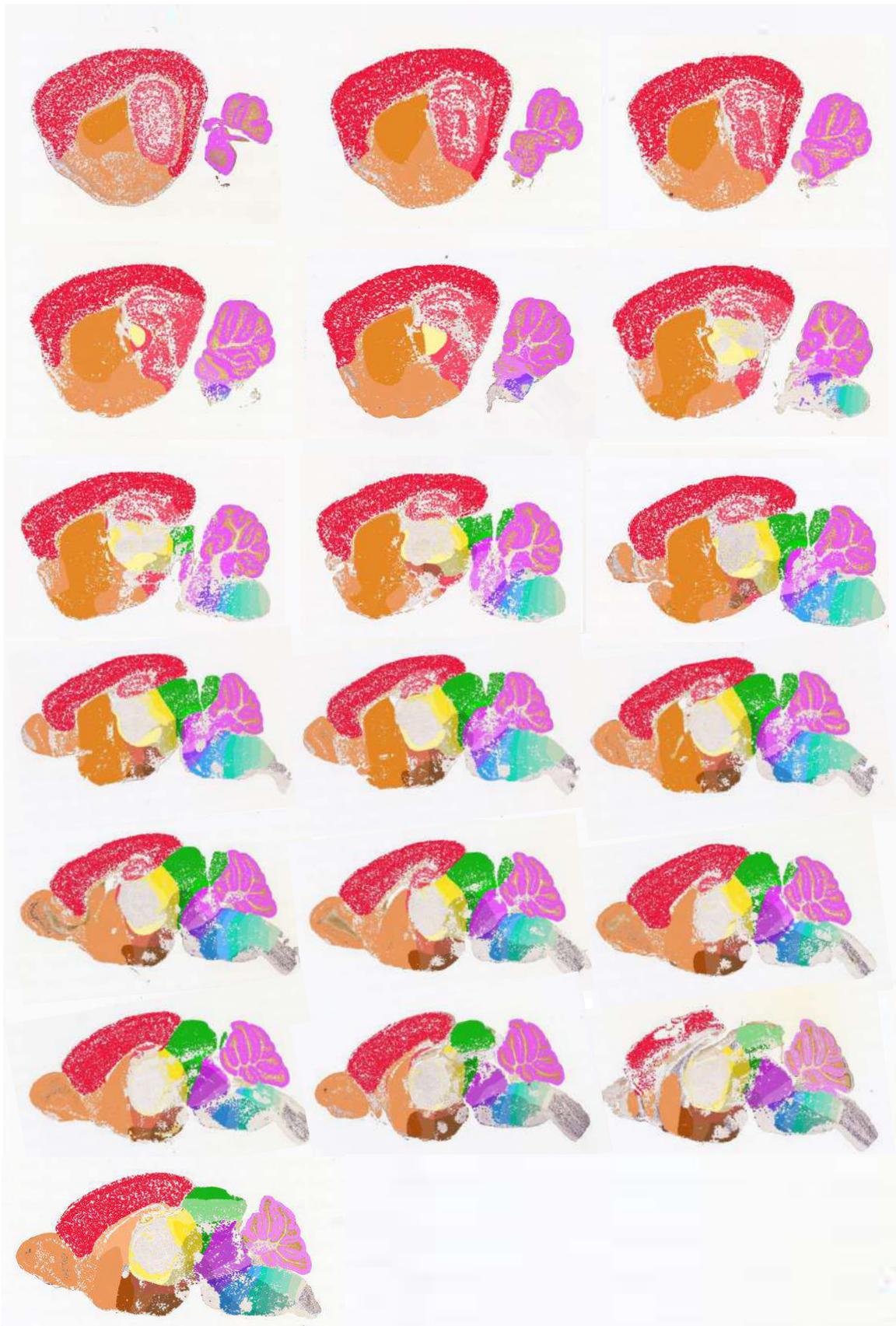

**Supplementary Figure 16. P14 VGAT brain Sections after passing through DeNeRD.**

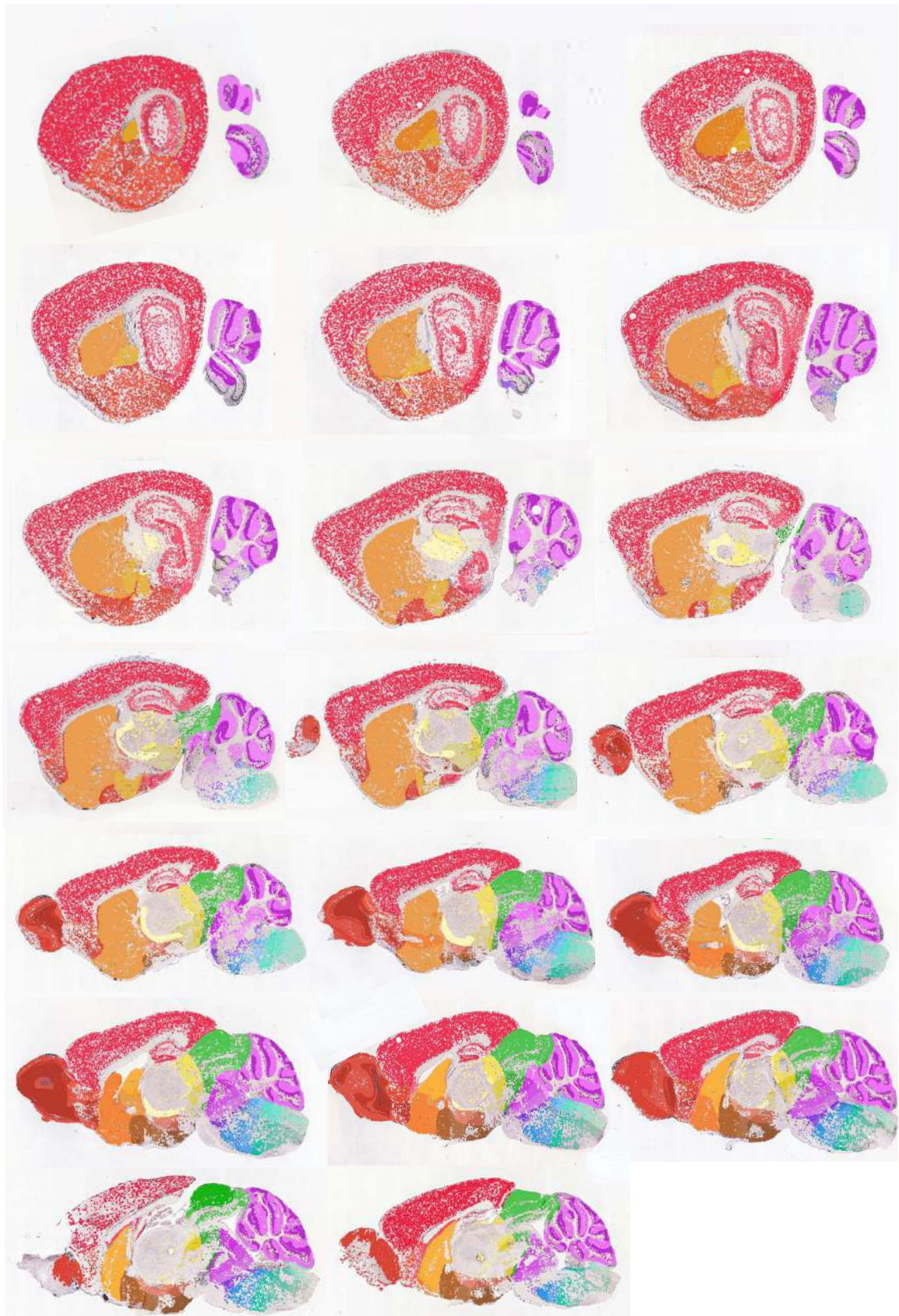

**Supplementary Figure 17. P56 GAD1 brain Sections after passing through DeNeRD.**

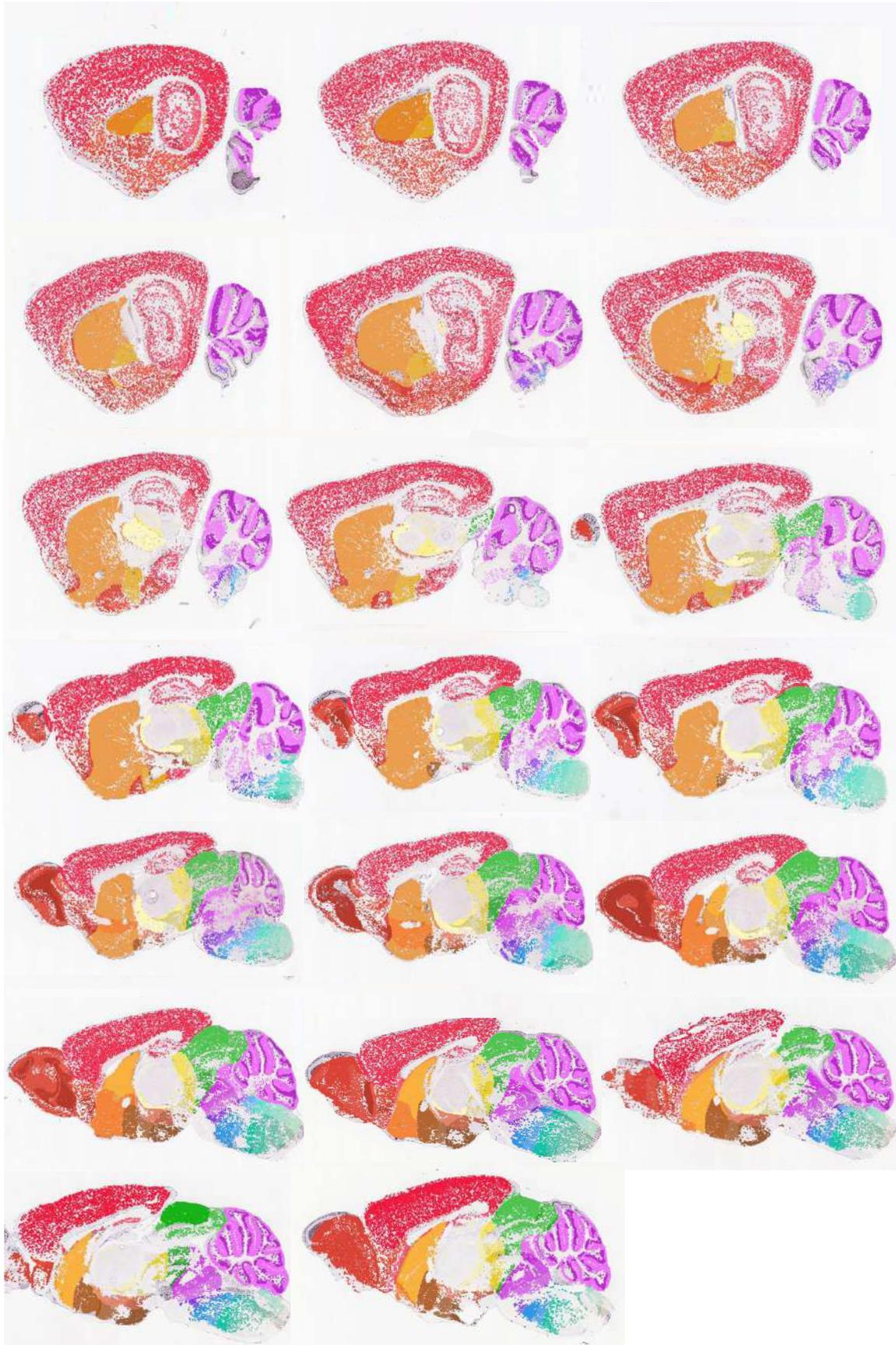

**Supplementary Figure 18. P56 VGAT brain sections after passing through DeNeRD.**

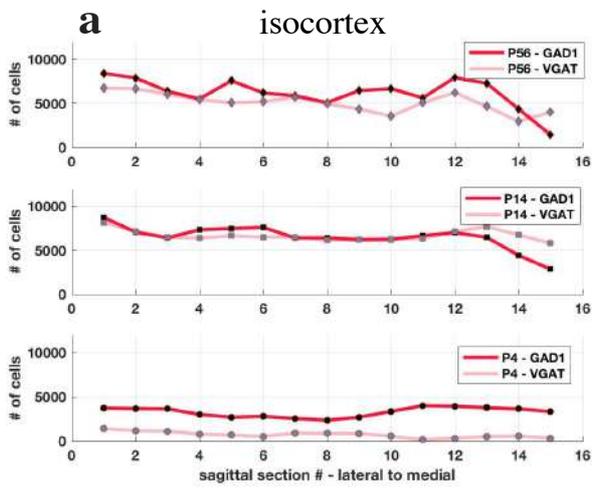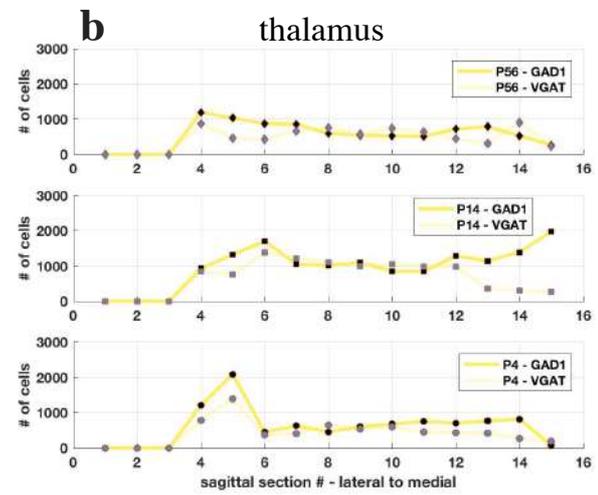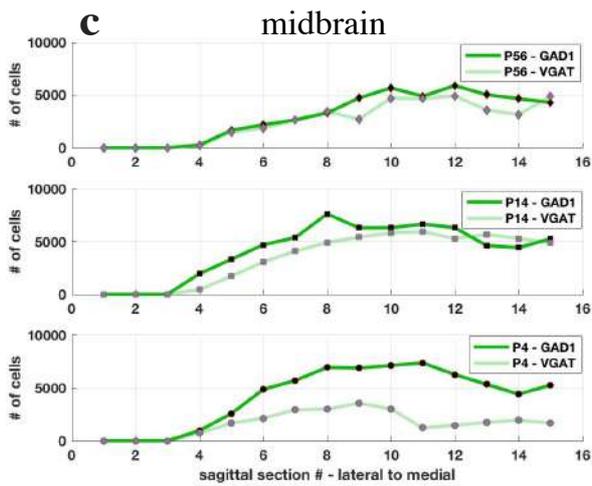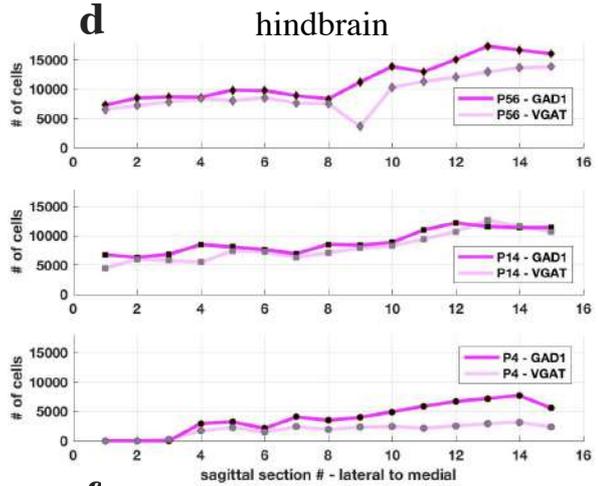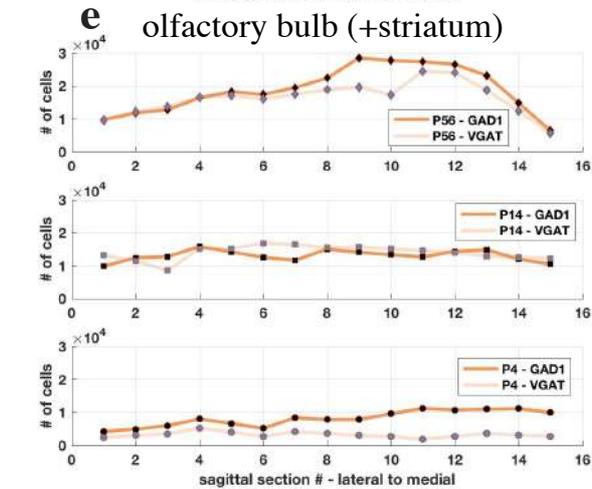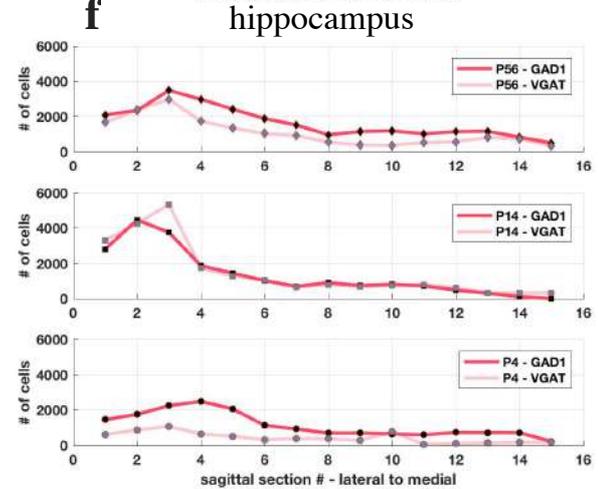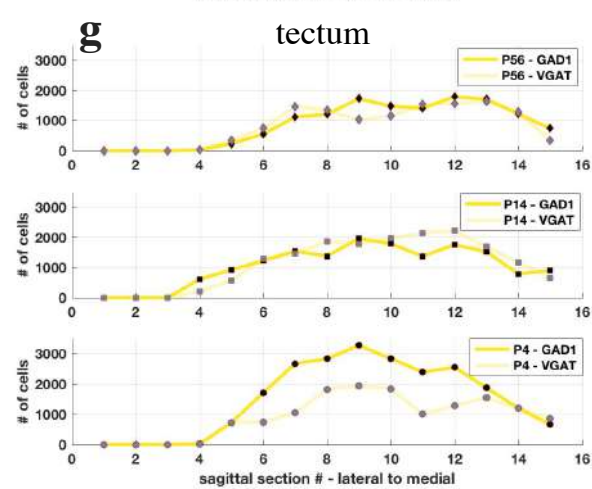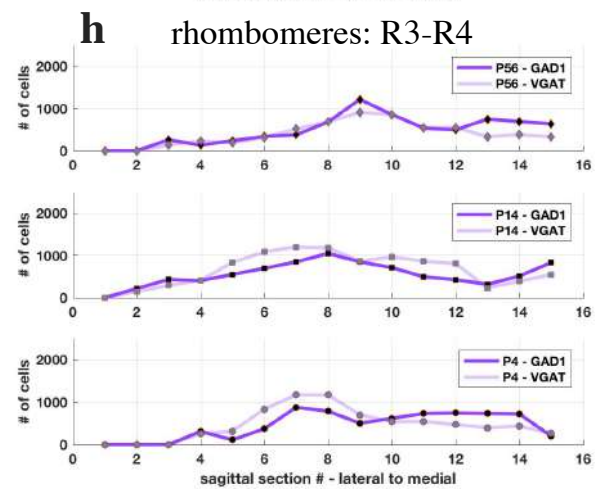

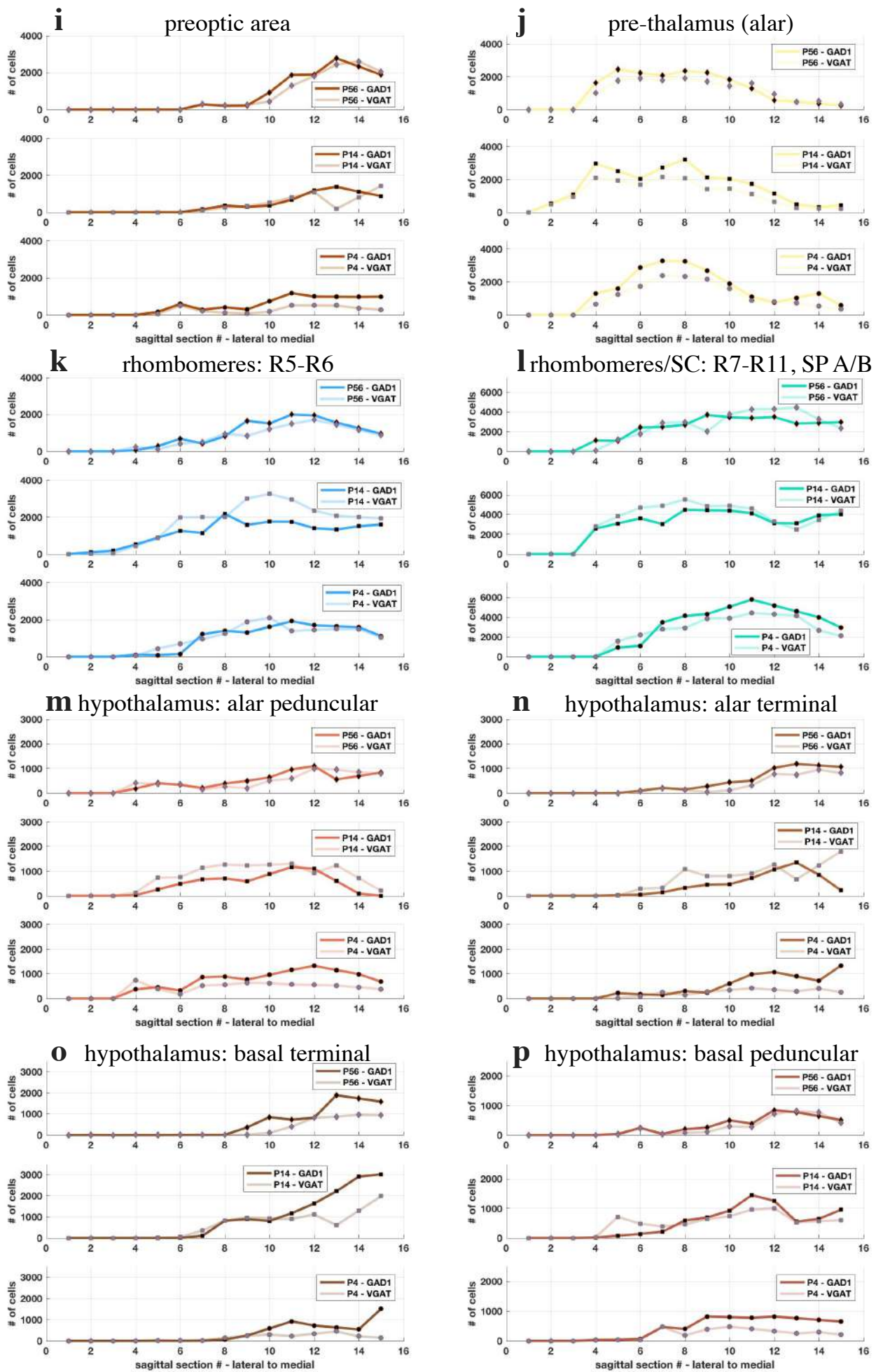

**Supplementary Figure 19.** Neural population in sagittal sections of GAD1 and VGAT mice brains at P4, P14 and P56 ages.